%% file: paper.tex
\documentclass[10pt,twocolumn,letterpaper]{article}

\usepackage{iccv}              % To produce the CAMERA-READY version

\usepackage{import}
\subimport{./}{packages}
\subimport{./}{macros}

\def\paperID{2839} % *** Enter the Paper ID here
\def\confName{ICCV}
\def\confYear{2025}

\title{\vspace{-2em} Synchronization of Multiple Videos}
\author{Avihai Naaman*\textsuperscript{\rm 1,2}, 
Ron Shapira Weber*\textsuperscript{\rm 1,2},
and
Oren Freifeld\textsuperscript{\rm 1,2,3}\\
$^{1}$Department of Computer Science, Ben-Gurion University of the Negev (BGU)\\
$^{2}$Data Science Research Center, BGU. \quad
$^{3}$School of Brain Sciences and Cognition, BGU.\\
}

\begin{document}
\maketitle

\begin{abstract}
Synchronizing videos captured simultaneously from multiple cameras in the same scene is often easy and typically requires only simple time shifts.
However, synchronizing videos from different scenes or, more recently, generative AI videos, poses a far more complex challenge due to diverse subjects, backgrounds, and nonlinear temporal misalignment. We propose Temporal Prototype Learning (TPL), a prototype-based framework that constructs a shared, compact 1D representation from high-dimensional embeddings extracted by any of various pretrained models. TPL robustly aligns videos by learning a unified prototype sequence that anchors key action phases, thereby avoiding exhaustive pairwise matching. Our experiments show that TPL improves synchronization accuracy, efficiency, and robustness across diverse datasets, including fine-grained frame retrieval and phase classification tasks. Importantly, TPL is the first approach to mitigate synchronization issues in multiple generative AI videos depicting the same action. 
Our code and a new multiple video synchronization dataset are available at \url{https://bgu-cs-vil.github.io/TPL/} 
%\vspace{-5mm}
\end{abstract}

\section{Introduction}\label{Sec:Intro}
\subimport{./paper}{introduction}
\section{Related Work}\label{Sec:Related}
\subimport{./paper}{related}
\section{Method}\label{Sec:Method}
\subimport{./paper}{method}
\section{Results}\label{Sec:Results}
\subimport{./paper}{results}

\section{Conclusion}\label{Sec:Conclusion}
\subimport{./paper}{conclusion}

\newpage
\clearpage
\section*{Acknowledgments}
This work was supported by the Lynn and William Frankel Center at BGU CS, 
by the Israeli Council for Higher Education via the BGU Data Science Research Center, and
by Israel Science Foundation Personal Grant \#360/21. R.S.W.'s work was supported by the Kreitman School of Advanced Graduate Studies. 

%\clearpage  % TODO REVIEW/FINAL: This \clearpage needs to be removed from both review and camera-ready versions.


{
    \small
    \bibliographystyle{ieeenat_fullname}
    \bibliography{refs.bib}
}

%%%%% SUPMAT %%%%%%%%%%%
\clearpage
\appendix
\onecolumn

% Manual supplementary title (no second \maketitle)
\begin{center}
{\LARGE \textbf{Synchronization of Multiple Videos}\\[0.4em]
\Large Supplementary Material}
\end{center}
\vspace{1em}

% If you want "Appendix A/B/..." numbering:
\setcounter{section}{0}
\renewcommand{\thesection}{\Alph{section}}

\section{GenAI Multiple Video Synchronization Dataset}
\subimport{./supmat}{dataset}
\section{Additional results}
\subimport{./supmat}{results}
\section{Implementation Details}
\subimport{./supmat}{training}

\end{document}

%% file: packages.tex
\usepackage{microtype}
\usepackage{graphicx} % only included once
\usepackage{booktabs} % for professional tables
\usepackage{xspace} % for et al./etc./e.g. macros
\usepackage{tikz}
\usetikzlibrary{matrix,positioning,calc}
\usetikzlibrary{arrows.meta}

\usepackage{multirow} 
\usepackage{pgfplots}

\usepackage{amsmath, amssymb, amsfonts} % AMS packages

\usepackage[thmmarks, amsmath, thref]{ntheorem} % Theorem environments
\usepackage{bbm}
\usepackage{bm}

\usepackage[normalem]{ulem} % for \sout without affecting \emph
\usepackage{pifont}% http://ctan.org/pkg/pifont
\newcommand{\cmark}{\ding{51}}  % ✓
\newcommand{\xmark}{\ding{55}}  % ✗
% Updated Times font package
\usepackage{newtxtext,newtxmath}

\usepackage{caption}
\usepackage{subcaption}
\usepackage{float}

% Algorithm package
\usepackage[linesnumbered, algoruled, noend, noline]{algorithm2e}

% Diagonal box in tables
\usepackage{diagbox}

\definecolor{iccvblue}{rgb}{0.21,0.49,0.74}
\definecolor{lightyellow}{rgb}{1.0, 1.0, 0.8} 

\usepackage[pagebackref,breaklinks,colorlinks,allcolors=iccvblue]{hyperref}
\usepackage{tikz}
\usetikzlibrary{shapes.geometric}

%% file: paper/introduction.tex
% % %%%%%%% FIGURE %%%%%%%%%%
\subimport{./}{fig_intro}
% % %%%%%%% FIGURE %%%%%%%%%%

%Synchronizing multiple videos that depict the same action is a core yet demanding task in computer vision. 
 Multiple Video Synchronization (MVS) of the same action is a challenging problem in computer vision, particularly in unconstrained settings. Standard solutions often rely on pairwise alignment~\cite{Debidatta:CVPR:2019:TCCL, Chen:CVPR:2022:CARL, zhang2023modeling, Liu:CVPR:2022:VAVA}, where each pair of videos is matched in isolation. Although such methods are relatively straightforward for small-scale problems, they suffer from two major shortcomings when extended to multiple videos.
The first is the high computational cost. Consider a dataset of $N$ videos, each containing $L$ frames, alongside a \emph{new} video of length $L$ for synchronization or frame retrieval. In a pairwise approach, every frame must be compared against all $N \times L$ frames in the training set, incurring an $O(N \times L^2)$ complexity. This exhaustive nearest-neighbor (NN) search is prohibitively expensive for real-world scenarios, where both $N$ and $L$ can be large.

The second is the lack of global consistency. Even if pairwise alignments yield accurate matches in isolation, they do not necessarily guarantee a \emph{joint} alignment across the entire collection of videos. Repeated pairwise matches can conflict, since different pairs may learn disparate references for similar action phases. As a result, there is no unified representation of the action progression that consistently aligns \emph{all} videos.

To address these issues, we advocate a \textbf{prototype-based} alignment strategy that bypasses the need for a single reference video and enables the synchronization of \emph{all} videos at once. We propose \textbf{Temporal Prototype Learning (TPL)}, which learns one-dimensional `bottleneck' signals capturing the underlying temporal structure (\ie., action prototypes) as universal anchors. By mapping each frame in every video to a shared temporal axis, TPL ensures global consistency and drastically reduces the computational cost. Synchronizing or retrieving a specific phase at time step $t$ for a new video thus amounts to referencing the $t$-th point in the learned prototype, rather than searching through the entire dataset.
\autoref{fig:intro} illustrates the TPL framework, where multiple videos are mapped to the same prototype space. 
Our main contributions are:
\begin{itemize}
    \item \textbf{Prototype-Based Synchronization:} We introduce a novel approach to jointly align multiple videos via a shared prototype space, overcoming the scalability and consistency challenges of pairwise methods.
    \item \textbf{Diffeomorphic Multitasking Autoencoder (D-MTAE)}: A novel architecture for learning one-dimensional `bottleneck' representation from multivariate video embeddings. D-MTAE can be trained on any pretrained video feature extractor and enables fast inference and robust multiple alignment. 
    \item \textbf{Linear-Time Frame Retrieval:} since, after alignment, semantically similar frames are mapped to the same time point,  frame retrieval simply entails returning all frames at that time point.
    \item \textbf{Synchronization of GenAI videos:} We show that TPL can sync not only multiple real-world videos but also multiple AI-generated videos depicting the same action. To demonstrate this, we also generated and annotated (for evaluation purposes only) the first GenAI-MVS dataset. 
    
\end{itemize}

%% file: paper/fig_intro.tex
\usetikzlibrary{positioning, arrows.meta, calc}
\usetikzlibrary{shapes.geometric}

\begin{figure*}[ht]
\centering
\resizebox{0.9\textwidth}{!}{
\begin{tikzpicture}[
  x={(1em,0)},y={(0,-1em)},
  box/.style={draw, minimum width=1cm, minimum height=4cm, align=center},
  arrow/.style={-{Latex[length=2mm]}, line width=0.8pt}
]

\definecolor{rongreen}{RGB}{44,160,44}
\definecolor{ronblue}{RGB}{31,119,180}
\definecolor{ronorange}{RGB}{255,127,14}
\definecolor{ronred}{RGB}{214,39,40}
\definecolor{ronmedpurp}{RGB}{147,112,219}
\definecolor{rontomato}{RGB}{255,99,71}
\definecolor{ronothergreen}{RGB}{2,103,63}
\definecolor{ronbrown}{RGB}{114,74,2}
\definecolor{ronotherblue}{RGB}{41,50,65}
\definecolor{ronotherorange}{RGB}{238,107,77}

%%%%% VIDS
\def\deltaframex{-2.5}
\def\deltaframey{-0.5}
\def\deltavidy{5}

\node (frozen) at (10,6.5) {\includegraphics[width=2.5em]{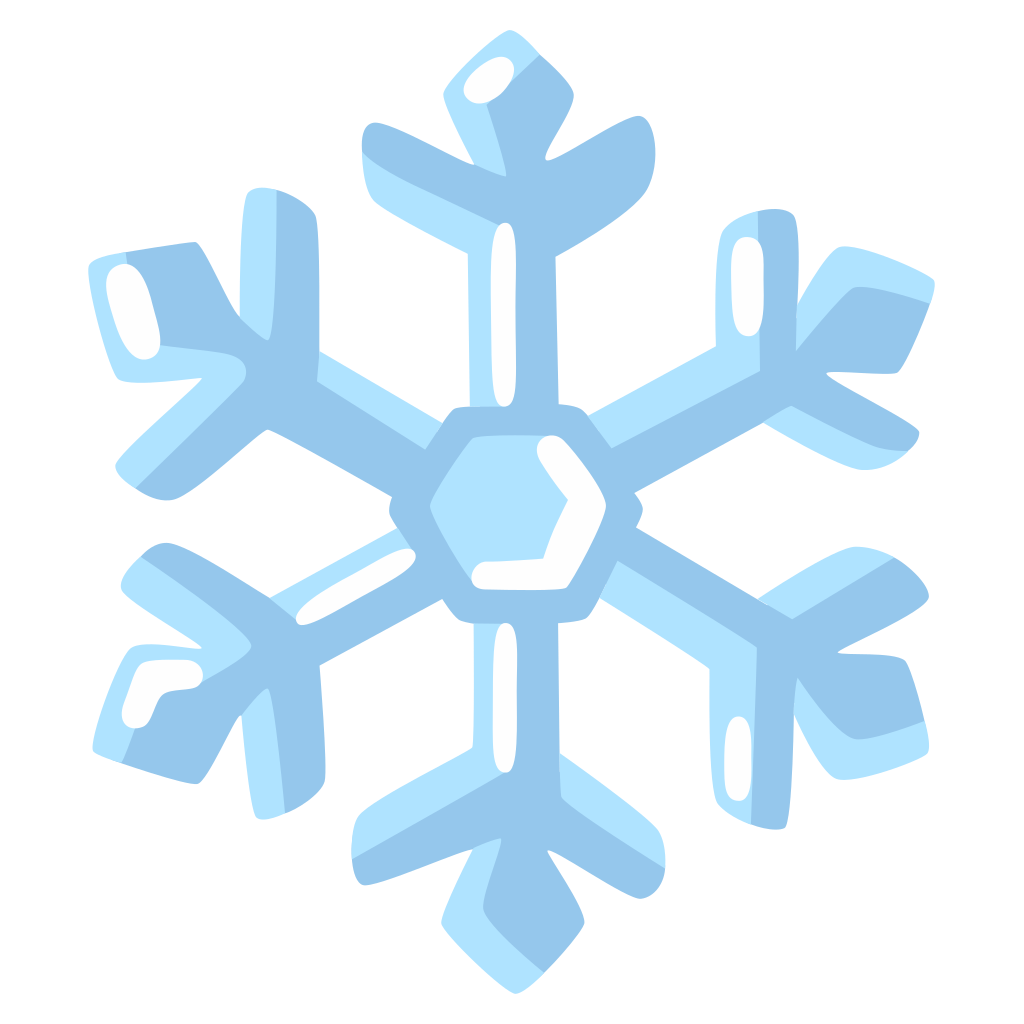}};

\node (v1f3) at (0+2*\deltaframex,-1*\deltavidy+2*\deltaframey) {\includegraphics[width=5em]{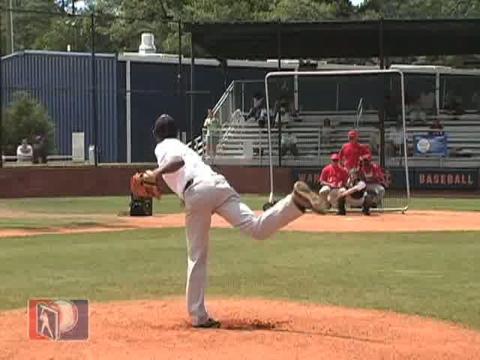}};
\node (v1f2) at (0+\deltaframex,-1*\deltavidy+\deltaframey) {\includegraphics[width=5em]{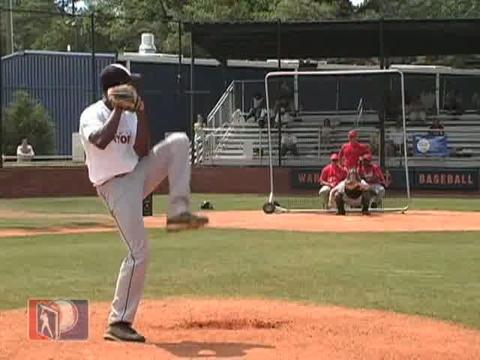}};
\node (v1f1) at (0,-1*\deltavidy) {\includegraphics[width=5em]{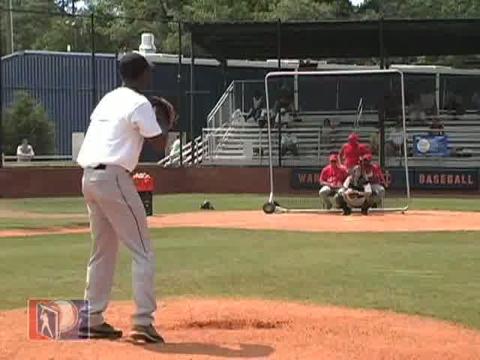}};
e
\node (v2f4) at (0+3*\deltaframex,0*\deltavidy+3*\deltaframey) {\includegraphics[width=5em]{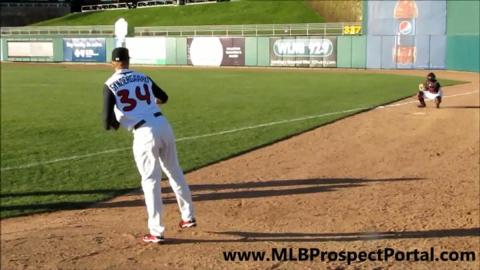}};
\node (v2f3) at (0+2*\deltaframex,0*\deltavidy+2*\deltaframey) {\includegraphics[width=5em]{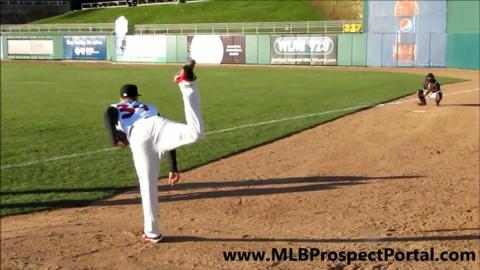}};
\node (v2f2) at (0+\deltaframex,0*\deltavidy+\deltaframey) {\includegraphics[width=5em]{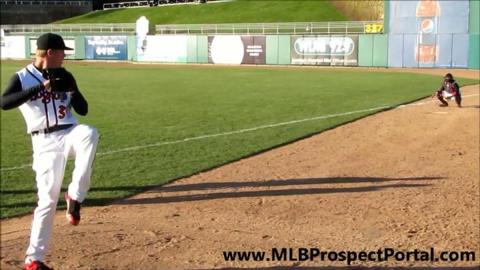}};
\node (v2f1) at (0,-0*\deltavidy) {\includegraphics[width=5em]{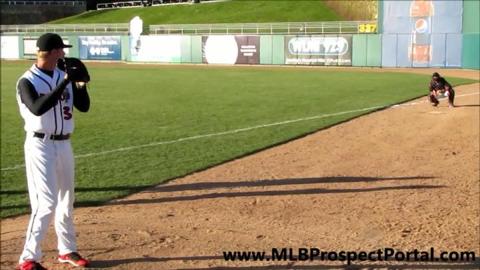}};

\node (v3f4) at (0+3*\deltaframex,1*\deltavidy+3*\deltaframey) {\includegraphics[width=5em]{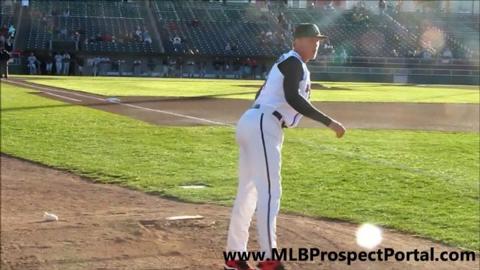}};
\node (v3f3) at (0+2*\deltaframex,1*\deltavidy+2*\deltaframey) {\includegraphics[width=5em]{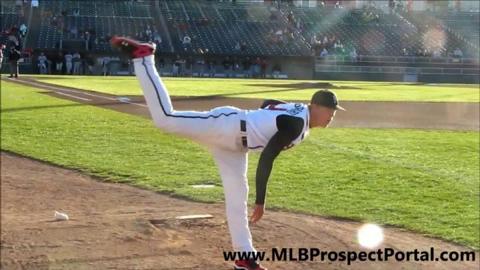}};
\node (v3f2) at (0+\deltaframex,1*\deltavidy+\deltaframey) {\includegraphics[width=5em]{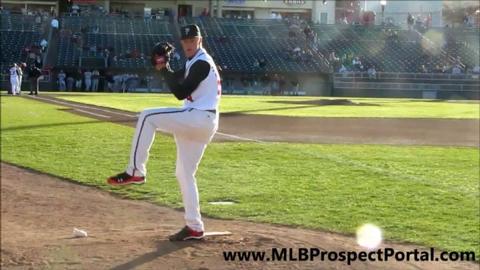}};
\node (v3f1) at (0,1*\deltavidy) {\includegraphics[width=5em]{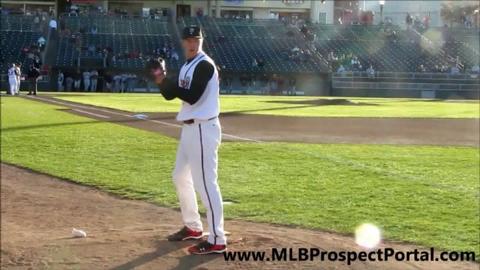}};

%\path (0,0*\deltavidy) -- (0,2*\deltavidy) node [font=\huge, midway, sloped] %{$\dots$};

\node (v4f5) at (0+4*\deltaframex,2*\deltavidy+4*\deltaframey) {\includegraphics[width=5em]{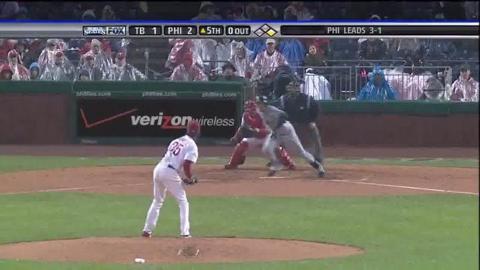}};
\node (v4f4) at (0+3*\deltaframex,2*\deltavidy+3*\deltaframey) {\includegraphics[width=5em]{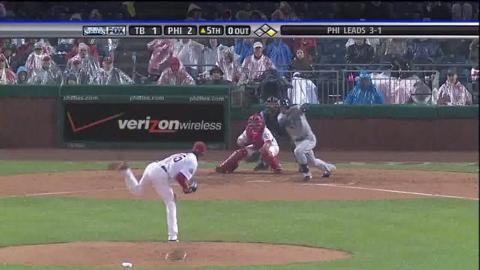}};
\node (v4f3) at (0+2*\deltaframex,2*\deltavidy+2*\deltaframey) {\includegraphics[width=5em]{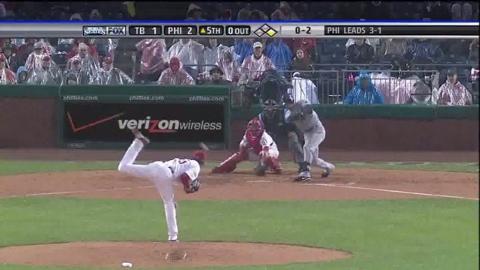}};
\node (v4f2) at (0+\deltaframex,2*\deltavidy+\deltaframey) {\includegraphics[width=5em]{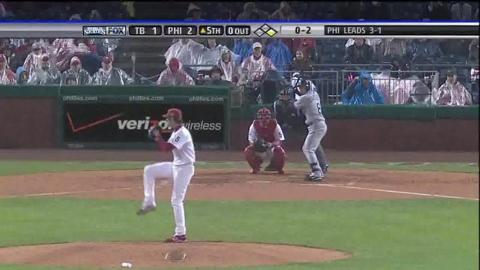}};
\node (v4f1) at (0,2*\deltavidy) {\includegraphics[width=5em]{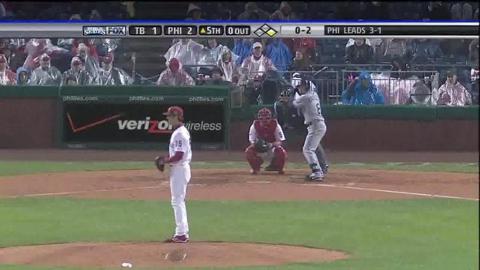}};

% \draw[-Stealth] (3,0) -- (7.75,0);
\draw[-Stealth, very thick, rongreen] (2.75,-1*\deltavidy) -- (5.5,-1*\deltavidy) -- (5.5,-1.5) -- (7.75,-1.5);
\draw[-Stealth, very thick, ronblue] (2.75,-0*\deltavidy) -- (4.5,-0*\deltavidy) -- (4.5,-0.5) -- (7.75,-0.5);
\draw[-Stealth, very thick, ronorange] (2.75,1*\deltavidy) -- (4.5,1*\deltavidy) -- (4.5,0.5) -- (7.75,0.5);
\draw[-Stealth, very thick, ronred] (2.75,2*\deltavidy) -- (5.5,2*\deltavidy) -- (5.5,1.5) -- (7.75,1.5);

%%%%% TCC
\node [trapezium, trapezium stretches=true, trapezium angle=60, minimum width=12em, minimum height=4em, draw, thick, shape border rotate=-180] (phi) at (10,0) {\huge$\phi$};

\draw[-Stealth, thick] (12.60,0) -- (15.5,0);

%%%%%
\def\deltachannelsy{3.6em}
\node[inner sep=0pt] (chn1) at (20,1.5*\deltachannelsy) {\includegraphics[width=10em]{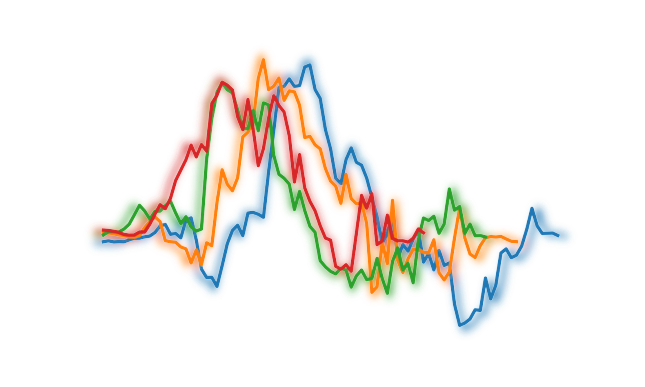}};

\node[inner sep=0pt] (chn2) at (20,0.5*\deltachannelsy) {\includegraphics[width=10em]{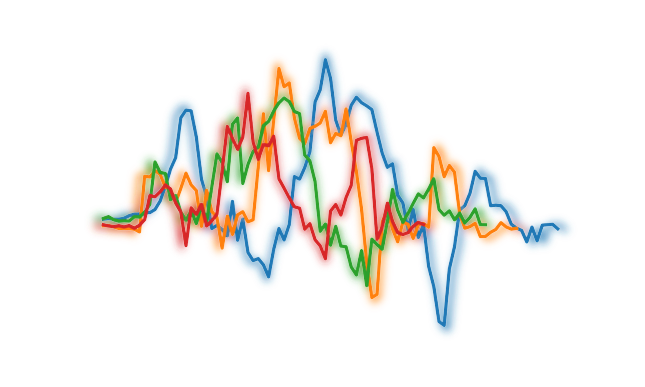}};

\path (19.5,0.5*\deltachannelsy) -- (19.5,-1.5*\deltachannelsy) node [font=\huge, midway, sloped] {$\dots$};

\node[inner sep=0pt] (chn3) at (20,-1.5*\deltachannelsy) {\includegraphics[width=10em]{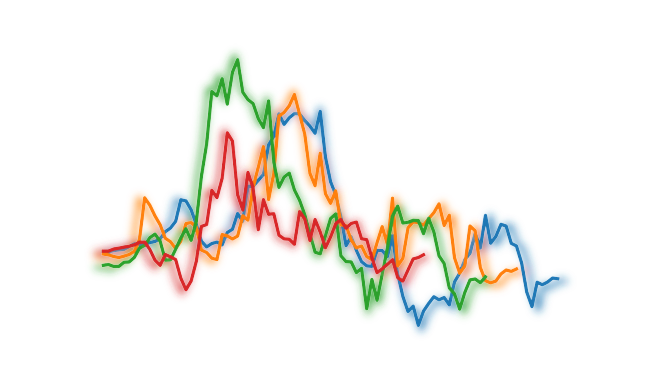}};

\node[above,font=\small] at (20,2.23*\deltachannelsy) {\Large Channels Representation};

\draw[-Stealth, thick] (24.0,0) -- (27.15,0);

% Fourth column
\node[box, minimum width=4em, 
    minimum height=14em, inner sep=8pt] (tpl) at (30,0) {\LARGE TPL};

\draw[-Stealth, thick] (32.85,0) -- (36.0,0);

% Fifth column
\node[inner sep=0pt] (achn1) at (40,1.5*\deltachannelsy) {\includegraphics[width=0.25\textwidth]{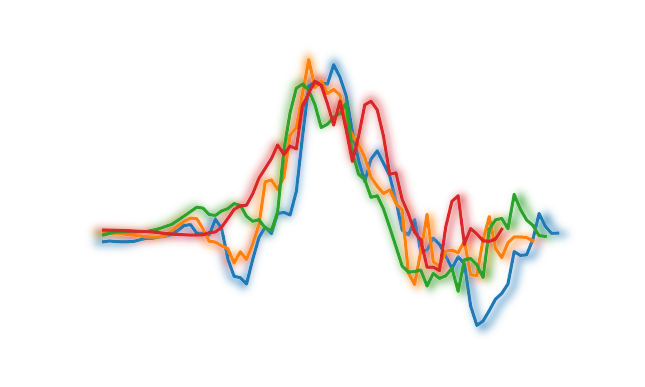}};

\node[inner sep=0pt] (achn2) at (40,0.5*\deltachannelsy) {\includegraphics[width=0.25\textwidth]{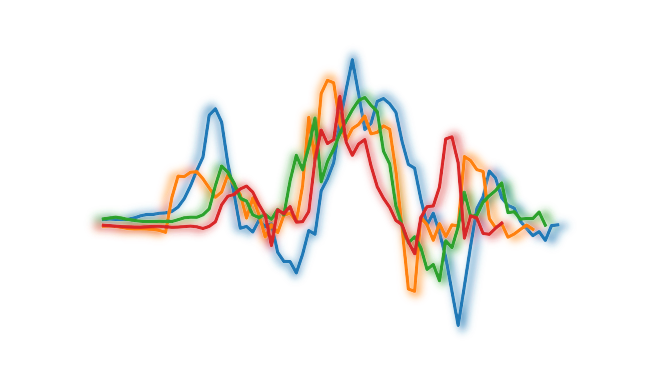}};

\path (40,0.5*\deltachannelsy) -- (40,-1.5*\deltachannelsy) node [font=\huge, midway, sloped] {$\dots$};

\node[inner sep=0pt] (achn3) at (40,-1.5*\deltachannelsy) {\includegraphics[width=0.25\textwidth]{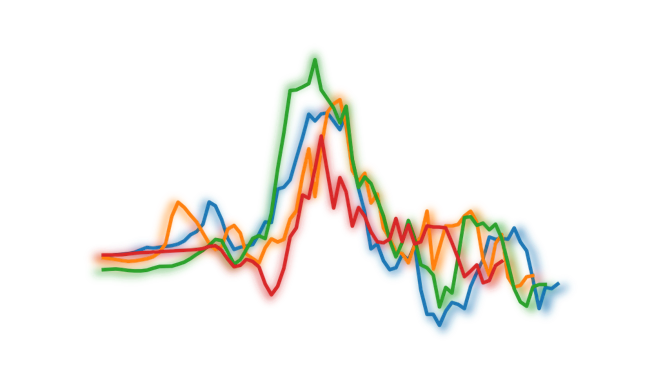}};

\node[above,font=\small] at (40,2.23*\deltachannelsy) {\Large Aligned};

% Sixth column
\node[inner sep=0pt] (achn1) at (50,1.5*\deltachannelsy) {\includegraphics[width=0.25\textwidth]{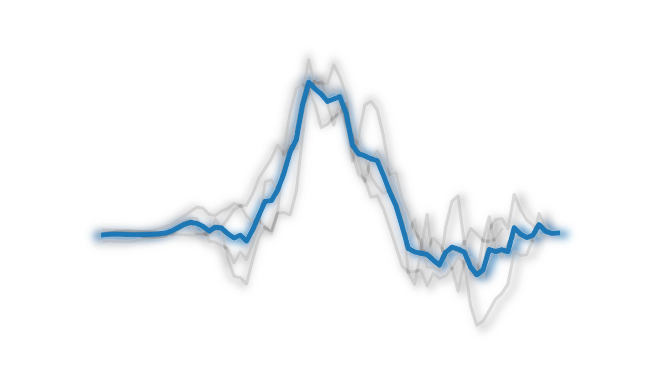}};

\node[inner sep=0pt] (achn2) at (50,0.5*\deltachannelsy) {\includegraphics[width=0.25\textwidth]{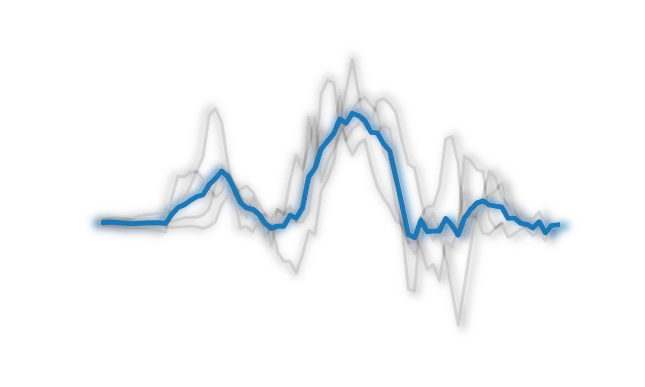}};

\path (50,0.5*\deltachannelsy) -- (50,-1.5*\deltachannelsy) node [font=\huge, midway, sloped] {$\dots$};

\node[inner sep=0pt] (achn3) at (50,-1.5*\deltachannelsy) {\includegraphics[width=0.25\textwidth]{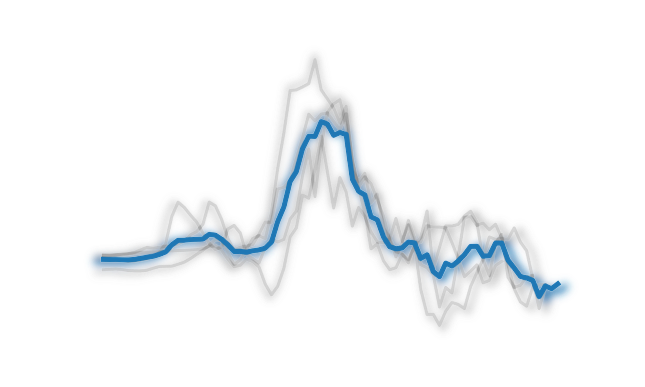}};

\node[above,font=\small] at (50,2.23*\deltachannelsy) { \Large Temporal Prototype};

% Key events misaligned

\draw[|-Stealth, ultra thick] (15,-3*\deltachannelsy) -- (25,-3*\deltachannelsy) node [above, pos=0.5] {Time};

\draw[|-Stealth, ultra thick] (35,-3*\deltachannelsy) -- (45,-3*\deltachannelsy) node [above, pos=0.5] {Time};

\node (exm1) at (15,-4.75*\deltachannelsy) {\includegraphics[trim=0 60 0 50, clip, width=9em]{figures/fig1/frames/a3.jpg}};

\node (exm2) at (25,-4.75*\deltachannelsy) {\includegraphics[trim=0 20 0 0, clip, width=9em]{figures/fig1/frames/b3.jpg}};

\node (exm3) at (35,-4.75*\deltachannelsy) {\includegraphics[trim=0 20 0 0, clip, width=9em]{figures/fig1/frames/c3.jpg}};

\node (exm4) at (45,-4.75*\deltachannelsy) {\includegraphics[trim=0 0 0 30, clip,width=9em]{figures/fig1/frames/d4.jpg}};

\draw[-, ronotherorange] (exm1.north) -- (19,-3*\deltachannelsy);
\filldraw[rongreen] (19,-3*\deltachannelsy) circle (2pt);
\draw[-, ronotherorange] (exm2.north) -- (23,-3*\deltachannelsy);
\filldraw[ronblue] (23,-3*\deltachannelsy) circle (2pt);
\draw[-, ronotherorange] (exm3.north) -- (17,-3*\deltachannelsy);
\filldraw[ronorange] (17,-3*\deltachannelsy) circle (2pt);
\draw[-, ronotherorange] (exm4.north) -- (21,-3*\deltachannelsy);
\filldraw[ronred] (21,-3*\deltachannelsy) circle (2pt);

\draw[-, ronotherblue] (exm1.north) -- (40,-3*\deltachannelsy);
\draw[-, ronotherblue] (exm2.north) -- (40,-3*\deltachannelsy);
\draw[-, ronotherblue] (exm3.north) -- (40,-3*\deltachannelsy);
\draw[-, ronotherblue] (exm4.north) -- (40,-3*\deltachannelsy);
\filldraw[ronotherblue] (40,-3*\deltachannelsy) circle (2pt);

\node[font=\large] at (30,-5.75*\deltachannelsy) {The \emph{Ball Release} key event};

\end{tikzpicture}
}
\captionof{figure}{\textbf{Temporal Prototype Learning (TPL)} uses an `off-the-shelf' feature extractor, denoted by $\phi$,
to generate initial multichannel action progression sequences for videos of the same action (\eg, Ball pitch~\cite{zhang:ICCV:2013:penn}).  Colors indicate different (and temporally-misaligned) videos of the same action. 
TPL produces the joint alignment and prototypical sequence, mapping key events (\eg, \emph{Ball Release})}
\label{fig:intro}
\end{figure*}

%% file: paper/related.tex
\paragraph{Video Representation Learning.}
Several approaches leverage sequence-level or pairwise matching signals to learn  representations from videos. TCC~\cite{Debidatta:CVPR:2019:TCCL} focuses on local alignment across pairs of videos, while GTA~\cite{Hadji:CVPR:2021:gta} extends this to longer sequences via a relaxed DTW-based contrastive loss. LAV~\cite{Haresh:CVPR:2021:LAV} introduces additional regularization to avoid trivial solutions, and VAVA~\cite{Liu:CVPR:2022:VAVA} allows variations in action order using priors on the optimal transport matrix. CARL~\cite{Chen:CVPR:2022:CARL} adopts a transformer-based contrastive framework with spatial and temporal augmentations. Although these methods have demonstrated success in pairwise settings, scaling them to large collections of videos typically requires extensive nearest-neighbor (NN) searches, which is computationally demanding and memory-intensive.
In parallel, large-scale pretrained image models such as DINO~\cite{Caron:ICCV:2021:dino} or OpenCLIP~\cite{Cherti:CVPR:2023:openclip} can be used for videos by extracting the \texttt{[CLS]} token or feature vector from each frame. Stacking these tokens over time yields a sequence of embeddings that capture both spatial semantics (from the pretrained image model) and temporal information (through the ordering of frames). We show TPL can use these embeddings for synchronizing multiple videos.   
\paragraph{Time-Series Alignment.}
Classical time-series alignment algorithms such as Dynamic Time Warping (DTW)~\cite{Sakoe:ICA:1971:DTW1,Sakoe:ASSP:1971:DTW2} and SoftDTW~\cite{Cuturi:2017:soft} are widely used to find optimal, order-preserving alignments between a pair of temporal sequences. SoftDTW, in particular, offers a differentiable variant, which has enabled end-to-end training when coupled with neural network feature extractors. However, these methods have a quadratic complexity in both time and memory \wrt{} sequence length, limiting their scalability. 
\paragraph{Prototype Learning and Temporal Prototypes.}
Prototype learning has proven effective in few-shot learning scenarios~\cite{Snell:NIPS:2017:prototypical}, where prototypical representations facilitate robust classification with minimal labeled data. In principle, a \emph{temporal} prototype can serve a similar role for video alignment, essentially summarizing the action progression into a single sequence. DTW Barycenter Averaging (DBA)~\cite{Petitjean:2011:global,Petitjean:2014:dynamic} and SoftDTW barycenters (SoftDBA)~\cite{Cuturi:2017:soft} offer ways to compute an average sequence under their respective distance function. Diffeomorphic Temporal Alignment Net (DTAN)~\cite{Shapira:NIPS:2019:DTAN,Martinez:ICML:2022:closed,Shapira:ICML:2023:rfdtan} learns diffeomorphic warping functions~\cite{Freifeld:ICCV:2015:CPAB,Freifeld:PAMI:2017:CPAB}, effectively jointly aligning all input sequences to their average. 
Although DTAN enables end-to-end learning of joint alignment (JA), it was not developed for aligning high-dimensional \emph{video embeddings} with large variation in length. In this work, we show how TPL addresses these issues. 

\paragraph{Multiple Video Synchronization: Limitations and Gaps.}
\label{sec:multi_video_limitations}
While pairwise alignment and small-scale joint alignment approaches have made significant progress, critical gaps remain for large-scale, \emph{multi-video} synchronization. First, naively extending pairwise methods to $N$ videos often leads to $O(N \times L^2)$ complexity in retrieval or synchronization, posing severe scalability constraints. Second, aligning pairs independently does not guarantee a \emph{globally-consistent} representation across all videos. Previous multiple video synchronization (MVS) methods focus on well-behaved scenarios that mainly involve multiple cameras recording the same scene, where the misalignment can be explained  
by a simple translation~\cite{Shrstha:ACM:2007:synchronization,Liang:ICASSP:2017:synchronization, Wang:ACM:2014:videosnapping}. However, synchronizing different scenes requires nonlinear alignment of the time axis. 

In this paper, we propose a new framework, called \emph{Temporal Prototype Learning (TPL)}, that addresses the limitations of existing approaches by jointly aligning multiple videos in a single, shared prototype space. This design enables robust synchronization, eliminates the need for exhaustive nearest-neighbor searches, and yields a linear-time retrieval mechanism for new video sequences.

%% file: paper/method.tex
We propose a novel approach for the synchronization of multiple videos without a reference.  
Our goal is to map similar action sequences to the same time step \wrt the action progression. To achieve this, we introduce TPL, which involves performing simultaneous dimensionality reduction and JA in the embedded space using a novel D-MTAE (\autoref{fig:arch1} depicts the framework). This section is organized as follows. We first review the required preliminaries in \autoref{subsec:method:preliminaries}. In~\autoref{Sec:Method:TPL}, we present a detailed explanation of the TPL framework, including its modules and loss functions.
In~\autoref{method:subsec:annotating}, we describe how to perform MVS and annotation transfer with TPL. Lastly, we discuss the limitations of TPL~\autoref{sec:method:subsec:limitation}.
%%%%%%%%%%
% FIGURE %
%%%%%%%%%%
\subimport{./}{fig_arch}
\subsection{Preliminaries}\label{subsec:method:preliminaries}
%%%%%%%%
% TABLE %
%%%%%%%%%
\subimport{tables}{tab_notations}
\paragraph{Notation and Setup (see~\autoref{tab:notations}).}
Consider $N$ videos, $(S_i)_{i=1}^N$. 
Let \(S_i = (s_{1}^i, s_{2}^i, \dots, s_{L}^i)\) be a video of length \(L\), where \(s_{t}^i\) is the \(t\)-th frame. We define the per-frame embedding $u_{t}^i = \phi\bigl(s_{t}^i)\in\mathbb{R}^C$, where $\phi$ is a feature extractor and $C$ is the number of channels (\ie, the embedding dimension) of the representation of the video.
The embedded feature sequence is $U_i=\{u_{t}^i\}_{t=1}^L \in \mathbb{R}^{C\times L}$. Thus, the set of video embeddings to be synchronized is $\{U_{i}\}_{i=1}^N$. $U_i$ could either be produced by applying an image-based classifier (\ie, the DINO \texttt{[CLS]} token~\cite{Caron:ICCV:2021:dino}) to each frame or a video-based one such as CARL~\cite{Chen:CVPR:2022:CARL}.
For each  $U_i$, we denote the predicted warping parameters and the corresponding time warp as $\btheta_i\in\RR^d$ and $T^{\btheta_i}$ respectively, such that $U_i\circ T^{\btheta_i}$ is the warped sequence
and $T^{\btheta_i}$
belongs to a $d$-dimensional parametric transformation family. 
 Finally, the average of
 the temporally-aligned sequences is $\widehat{U}=\frac{1}{N}\sum\nolimits_{i=1}^NU_i\circ T^{\btheta_i}$.

The Joint Alignment (JA) problem can then be thought of as finding 
the set of warping parameters between $\{U_{i}\}_{i=1}^N$ and $\widehat{U}$ which minimize their discrepancy, $D$ (\eg, the Euclidean distance). Since $\widehat{U}$ is unknown, the JA problem becomes: 
\begin{align}\label{eq:JA}
    (T^{\btheta_i^{*}})_{i=1}^N,\mu   = \argmin{(T^{\btheta_i})_{i=1}^N\in\Tcal, U } \sum\limits_{i=1}^N 
     D(U , U_i\circ T_i)\,    
\end{align}
where $(T^{\btheta_i^{*}})_{i=1}^N$ and $\mu$ denote the optimal warping parameters and average sequence, respectively, and $\Tcal$ is the transformation family (\eg, phase-shift, elastic, \etc). Partly due to the unsupervised nature of the task, a regularization term is usually added to avoid trivial solutions and/or unrealistic deformations. The problem is then reformulated as: 
\begin{align}\label{eq:JA:reg}
    (T^{\btheta_i^{*}})_{i=1}^N,\mu   = \argmin{(T^{\btheta_i})_{i=1}^N\in\Tcal, U } \sum\limits_{i=1}^N 
     D(U , U_i\circ T_i)+ \Rcal(T^{\btheta_i};\lambda) \,    
\end{align}
where $\Rcal(T^{\btheta_i};\lambda)$ is the regularizer over $T^{\btheta_i}$, and $\lambda$ is a hyperparameter (HP) controlling the regularization strength. An important, yet often overlooked, fact is that $\lambda$ is usually dataset specific, must be found via an expensive search, and that finding a good value requires
supervision (\ie, ground-truth
labels are needed to rank the performance with different values of $\lambda$). To alleviate this issue, we follow
a regularization-free approach~\cite{Shapira:ICML:2023:rfdtan} to JA which uses the Inverse-Consistency Averaging Error  (ICAE; detailed below), 

\paragraph{Diffeomorphic Temporal Alignment Nets (DTAN).}
DTAN~\cite{Shapira:NIPS:2019:DTAN,Martinez:ICML:2022:closed,Shapira:ICML:2023:rfdtan} is a learning-based model designed for time series JA. Given $N$ sequences $\{U_{i}\}_{i=1}^N$, DTAN predicts a set of continuous time-warp parameters $\{\btheta_{i}\}_{i=1}^N$ to minimize the within-class variance. This is %analogous
akin to finding the average sequence. 
These warps are applied via CPAB transformations~\cite{Freifeld:ICCV:2015:CPAB,Freifeld:PAMI:2017:CPAB} (described below). 
DTAN has been designed for \emph{univariate} time series and was evaluated on the relatively `well-behaved' UCR archive~\cite{Dau:2019:ucr}. While DTAN could arguably be generalized to multivariate representations of videos, applying a single warp to each multivariate sequence can overlook channel-specific temporal variations that, in turn, hinder the average sequence's computation.
Another limitation specific to  ICAE is that the JA of variable-length multivariate data
(as opposed to variable-length univariate data) usually results in a `shrinking' effect, where the average sequence length is much shorter than the original data. 

Our proposed TPL resolves these issues by 1) introducing a univariate “bottleneck” that discards channel-specific variations not shared across all sequences, and 2) setting the average sequence to match the median length of the data. This design allows for robust JA in the high-dimensional embedding space while retaining the desirable properties of DTAN. That is, end-to-end, misalignment-invariant learning of a shared temporal structure across multiple videos. 

\paragraph{CPAB Transformations.}
The CPAB (Continuous Piecewise Affine-Based) warp~\cite{Freifeld:ICCV:2015:CPAB,Freifeld:PAMI:2017:CPAB} lies at the core of DTAN. Unlike discrete alignment approaches (\eg, DTW), a CPAB transformation is parameterized by a parameter vector \(\btheta\) that defines a Continuous Piecewise Affine (CPA) velocity field, $\bv^\btheta$, such that its integration yields a diffeomorphism
(namely, a smooth, differentiable map, with a differentiable inverse), $T^\btheta$. In the context of time series, this is a differentiable order-preserving time warp. 
This approach has three major advantages for learning-based alignment:
\begin{enumerate}
    \item \textbf{Efficiency and Accuracy:} CPA velocity fields permit fast and accurate integration~\cite{Freifeld:ICCV:2015:CPAB,Freifeld:PAMI:2017:CPAB}, making them suitable for large-scale video data.
    \item \textbf{Closed-Form Gradients:} The CPAB gradient, $\nabla_\btheta T^{\btheta}$, also admits a closed-form solution~\cite{Martinez:ICML:2022:closed}, which enables stable end-to-end training of neural alignment models.
    \item  \textbf{Invertibility and symmetry:} CPAB warps are invertible, where $(T^{\btheta})^{-1}=T^{-\btheta}$. This is in contrast to DTW, which might produce different warping paths for $\textrm{DTW(X,Y)}$ and $\textrm{DTW(Y,X)}$, 
\end{enumerate}
Once a DTAN has been trained for a particular class of sequences, it can be applied directly to new data without re-solving an alignment objective from scratch, thus making the entire pipeline efficient for both training and inference.

\subsection{Temporal Prototype Learning}\label{Sec:Method:TPL}
\paragraph{Architecture.}
Given $N$ videos depicting the same action and their high-dimensional embeddings, $\{U_i\}_{i=1}^{N}$, we seek to learn a temporal prototype,  $\widehat{U}\in\RR^{C\times L}$ where $L$ is the prototype's length and $C$ is the number of channels in the learned representation. Since $\{U_i\}_{i=1}^{N}$ are misaligned, a simple averaging will result in a distorted average sequence that represents the data
poorly. Another key insight is that while at each time $t$, $u_t^i\in\RR^{C}$, the $(u_t^i)_{t=1}^{L_i}$ values (where $L_i$ is the length of $S_i$, hence also of $U_i$) should represent, in theory, phases in a 1D action progression. Thus, to learn temporal prototypes of action progression, a 1D representation should suffice. This is further motivated by the fact that the high-dimensional representation might hold irrelevant information, which hinders the alignment task. Taking the discussion above into consideration, we propose a \emph{simultaneous dimensionality reduction and JA} to achieve a compact representation of the action and its progression.

Specifically, we introduce a novel Diffeomorphic Multitasking Autoencoder (D-MTAE; depicted in~\autoref{fig:arch1}) designed to learn dimensionality reduction and joint alignment. 
D-MTAE consists of: 1) an encoder, $\Psi_{\text{encoder}}:\RR^{C\times L_i}\to\RR^{L_i}$, that maps the $C$-dimensional embedding sequence, $U_i\in\RR^{C\times L_i}$, into a latent 1D projection, $Z_i\in\RR^{L_i}$; 
2) 
an alignment module, 
$\Psi_{\text{Align}}$, that performs JA on the latent representations, $(Z_i)_{i=1}^N$;
3) a decoder model, $\Psi_{\text{decoder}}:\RR^{L_i}\to\RR^{C\times L_i}$, that maps the latent projection back to the original domain. 

\paragraph{Latent Representation Alignment Loss.}
The encoder, $\Psi_{\text{encoder}}$, is a Temporal Convolutional Network (TCN) that maps each $U_i$ to a univariate latent sequence $Z_i \in \RR^{L_i}$. The alignment module $\Psi_{\text{Align}}$ predicts warping parameters $\{\btheta_i\}_{i=1}^{N}$ to produce time-warped latent signals $\widetilde{Z}_i = Z_i \circ T^{\btheta_i}$. We seek a shared prototype $\widehat{Z} \in \RR^{L}$ that captures the common temporal progression across all videos. Building on the Inverse Consistency Averaging Error (ICAE)~\cite{Shapira:ICML:2023:rfdtan}, which enables JA without explicit warp regularization, we minimize
\begin{align}
\label{eq:icae:z}
\Lcal_{\mathrm{ICAE}} 
\;=\;
\frac{1}{N}\sum_{i=1}^{N}
\Bigl\|
\,\widehat{Z} \circ T^{-\btheta_i} - Z_i
\Bigr\|_{\ell_2}^{2}.
\end{align}
Since videos can vary greatly in length, we fix the length of $\widehat{Z}$ to be the \emph{median} of all video lengths to prevent ``collapse'' of the prototype (observed empirically when video lengths differ significantly). This approach robustly maintains an appropriate temporal scale in the aligned representation.

\paragraph{Misalignment-Invariant Reconstruction Loss.}
Ensuring that $\widehat{Z}$ accurately reflects the data’s true progression requires preventing trivial solutions (\eg, collapsing each $\widetilde{Z}_i$ to a single repeated scalar). To address this, we include a decoder $\Psi_{\text{decoder}}$ that reconstructs the original embeddings from the aligned latents, yielding $\widetilde{U}_i = \Psi_{\text{decoder}}(\widetilde{Z}_i)$. We then apply the inverse warp $T^{-\btheta_i}$ to $\widetilde{U}_i$ and measure the discrepancy from the original embeddings $U_i$:
\begin{align}
\label{eq:rec}
\Lcal_{\text{rec}}
=
\frac{1}{N}\sum_{i=1}^{N}
\Bigl\|
U_i
- 
\widetilde{U}_i \circ T^{-\btheta_i}
\Bigr\|_{\ell_2}^{2}.
\end{align}
This \emph{misalignment-invariant} reconstruction encourages the prototype to capture meaningful temporal structure, as it must remain consistent when warped back to each video’s original timeline.

\textbf{Overall loss:} The overall loss function is obtained by combining the JA loss (\autoref{eq:icae:z}) and reconstruction loss (\autoref{eq:rec})
\begin{align}\label{eq:full}
   \Lcal_{\text{TPL}} = \lambda_t \Lcal_{\text{ICAE}} + \Lcal_{\text{rec}}\,
\end{align}
where $\lambda_t$ controls the `annealing' of the alignment loss. This allows for faster and reliable convergence for the simultaneous learning of reconstruction and alignment. It is defined as
\begin{align}\label{eq:annealing}
 \lambda_t = \frac{1}{1 + e^{-\alpha (t - t_0)}} \,
\end{align}
where $\alpha$ is a scaling factor (fixed at $2$ for all experiments),  $t$ is the current training epoch, and $t_0$ is the epoch at which \( \lambda_t \) reaches 1 and the annealing stops (set to $\frac{N_{\textrm{epochs}}}{2}$).

The D-MTAE is trained simultaneously in an end-to-end fashion.  We used PyTorch~\cite{Paszke:NIPS:2019:pytorch} for all of our experiments. The \texttt{DIFW} package~\cite{Martinez:ICML:2022:closed} was used for the CPAB~\cite{Freifeld:ICCV:2015:CPAB,Freifeld:PAMI:2017:CPAB} implementation. Both $\Psi_{\text{encoder}}$ and $\Psi_{\text{decoder}}$ are 3-layer TCNs. For $\Psi_{\text{Align}}$ we follow~\cite{Shapira:ICML:2023:rfdtan} and use InceptionTime~\cite{Ismail:2020:inceptiontime}. 
For a detailed description of the training procedure and hyperparameters, see our supplementary material (\textbf{SupMat}).
%%%%% FIGURE %%%%%%%%
\subimport{./}{fig_joint_alignment}
%%%%% FIGURE %%%%%%%%
%%%%%%%%%%
% FIGURE %
\subimport{../}{paper/fig_genai.tex}
%%%%%%%%%%
\subsection{Multiple Video Synchronization}\label{method:subsec:annotating}
Aligning new videos is achieved by 
first predicting the warping parameters for the latent representations, $(\btheta_i)_{i=1}^{N}$, and applying them to the original videos; \ie, $(S_i\circ T^{\btheta_i})_{i=1}^N$. 
 The temporal prototype is defined as the average of the representation of the aligned sequences:
\begin{align}\label{eq:tpl:average}
\widehat{U}\triangleq 
 \tfrac{1}{N}\sum\nolimits_{i=1} U_i\circ T^{\btheta_i}\, .
\end{align}
\\
 Once the temporal prototypes are computed, we can transfer dense, frame-level, annotations from training to test data. This is achieved by first annotating the prototypes and then transferring their annotations to the test videos using temporal alignment. 
 Formally, let $(A_i)_{i=1}^{N}$ be the dense annotations of the input videos (\ie, $A_i$ is a sequence of length $L_i$ where $A_i[t]$ is the frame label at time step $t$). 
 To annotate the temporal prototype, we take the \text{mode}
 (\ie, the most frequent label) of the \emph{aligned} annotations at time step $t$,  $(A_i\circ T^{\theta_i})[t]$. The prototype labels, $\widehat{A}$, for each time step are defined as
\begin{align}
    \widehat{A}[t]\triangleq \text{mode}(\Acal[t])
\end{align}
where $\Acal[t]=((A_i\circ T^{\theta_i})[t])_{i=1}^{N}$ are all labels at time step $t$ after alignment.
New videos are annotated by aligning them to their corresponding class prototype and using the matching frame labels. ~\autoref{fig:JA:1D} shows the the 1D representations colored by the ground-truth annotations of a 20 \emph{Baseball swing} videos~\cite{zhang:ICCV:2013:penn} before and after synchronization.

\subsection{Limitations}\label{sec:method:subsec:limitation}
TPL's effectiveness is intricately tied to the quality of the initial features, \ie, the initial embeddings used. Should these embeddings be of poor quality or fail to adequately represent the data, the resulting outcomes may be suboptimal.

%% file: paper/fig_arch.tex
\usetikzlibrary{positioning, arrows.meta, calc}
\usetikzlibrary{shapes.geometric}
\usetikzlibrary{backgrounds,fit}

\definecolor{rongreen}{RGB}{44,160,44}
\definecolor{ronblue}{RGB}{31,119,180}

\begin{figure*}[t]
\centering
\resizebox{0.95\textwidth}{!}{
\begin{tikzpicture}[
  x={(1em,0)},y={(0,-1em)},
  box/.style={draw, minimum width=1cm, minimum height=4cm, align=center},
  arrow/.style={-{Latex[length=2mm]}, line width=0.8pt}
]

\definecolor{rongreen}{RGB}{44,160,44}
\definecolor{ronblue}{RGB}{31,119,180}
\definecolor{ronorange}{RGB}{255,127,14}
\definecolor{ronred}{RGB}{214,39,40}
\definecolor{ronmedpurp}{RGB}{147,112,219}
\definecolor{rontomato}{RGB}{255,99,71}
\definecolor{ronothergreen}{RGB}{2,103,63}
\definecolor{ronbrown}{RGB}{114,74,2}
\definecolor{ronotherblue}{RGB}{41,50,65}
\definecolor{ronotherorange}{RGB}{238,107,77}

\def\deltacol{8}

%%%%% PHI
\node (frozen) at (-1,3) {\includegraphics[width=2.5em]{paper/frozen.png}};

\node [trapezium, fill=white, trapezium stretches=true, trapezium angle=65, minimum width=12em, minimum height=5em, draw, thick, shape border rotate=-180] (phi) at (-2.5,-2.5) {\huge$\phi$};

\draw[-Stealth, very thick] (0.2,-2.5) -- (0.95*\deltacol-4.5,-2.5);

%%%%% CHN

\def\deltachannelsy{4em}
\node[inner sep=0pt] (chn1) at (0.8*\deltacol,3*\deltachannelsy) {\includegraphics[width=10em]{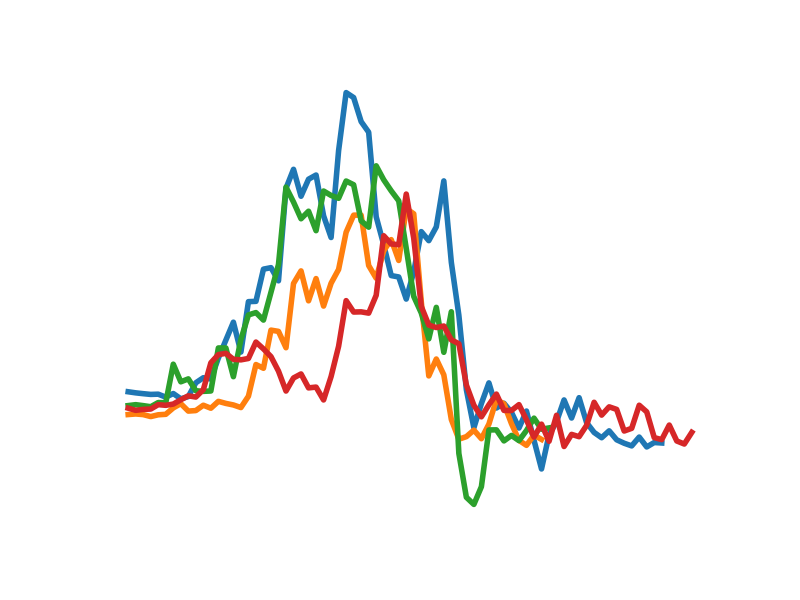}};

\node[inner sep=0pt] (chn2) at (0.8*\deltacol,1.8*\deltachannelsy) {\includegraphics[width=10em]{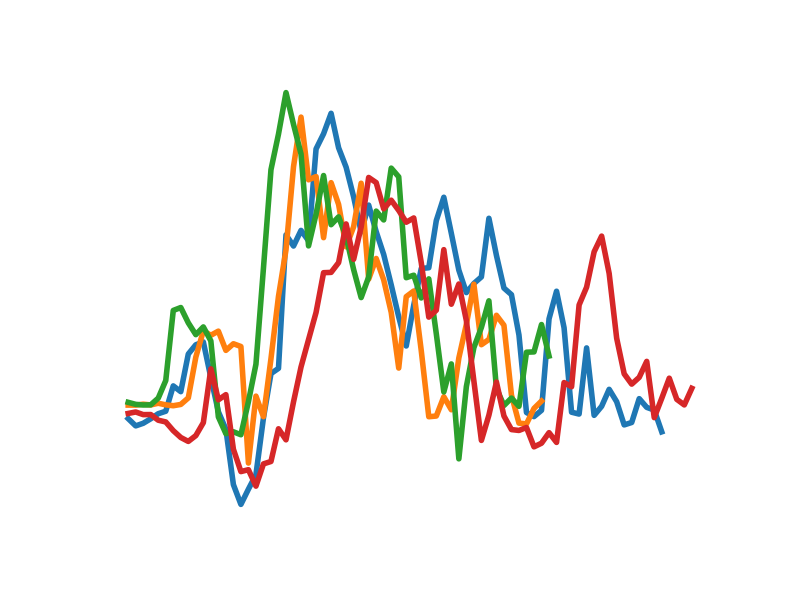}};

\path (0.75*\deltacol,0.35*\deltachannelsy) -- (0.75*\deltacol,0.35*\deltachannelsy) node [font=\huge, midway, sloped] {$\dots$};

\node[inner sep=0pt] (chn3) at (0.8*\deltacol,-1.2*\deltachannelsy) {\includegraphics[width=10em]{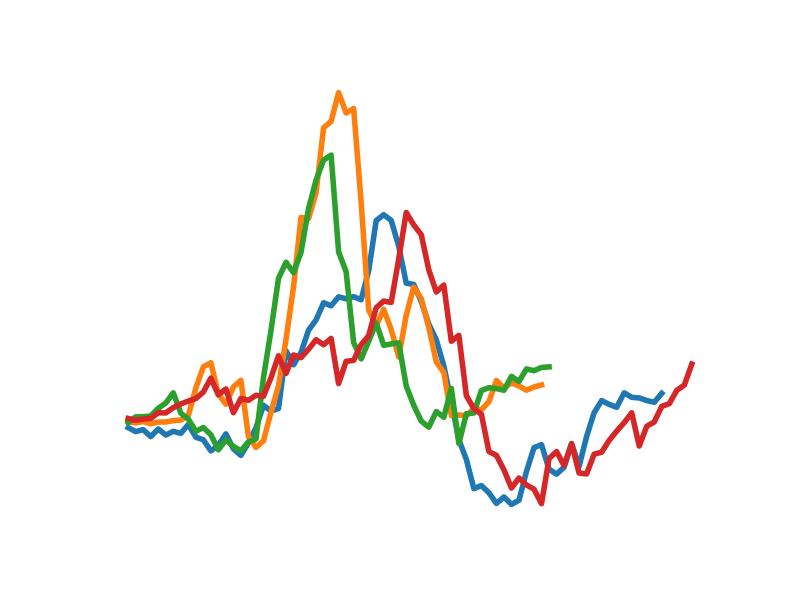}};

\node[above,font=\small] (top1) at (0.8*\deltacol,3.7*\deltachannelsy) {\large $(U_i)^{N}_{i=1}\in \mathbf{R}^{C \times L_i }$};

%%%% PHI ENC

\node [trapezium, fill=white, trapezium stretches=true, trapezium angle=80, minimum width=12em, minimum height=5em, draw, thick, shape border rotate=-180] (phienc) at (2*\deltacol,0) {\LARGE$\Psi_\mathrm{enc}$};

\draw[-Stealth, very thick] (2*\deltacol+2.75,0) -- (3*\deltacol-3,0);

%%%% CHN2

\node[inner sep=0pt] at (3*\deltacol,0) {\includegraphics[width=8em]{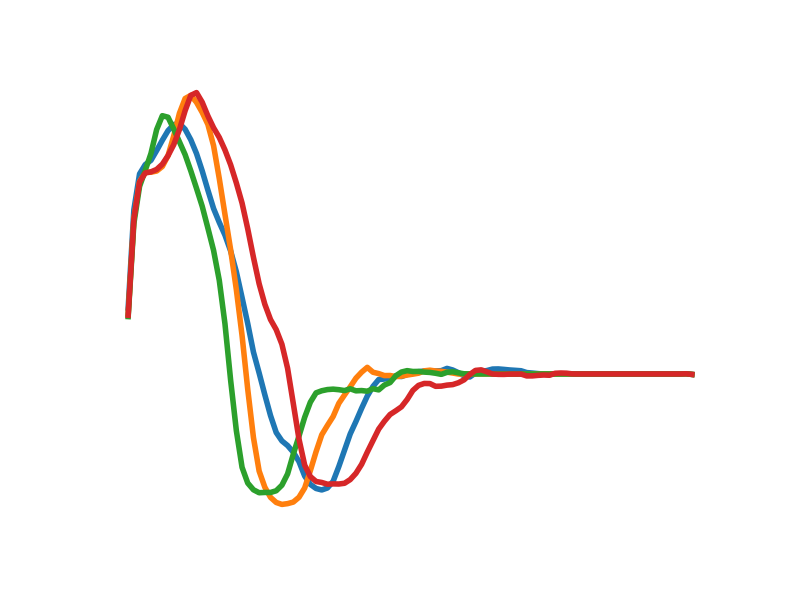}};

\node[above,font=\small] (top2) at (3*\deltacol,0.6*\deltachannelsy) {\large$(Z_i)^{N}_{i=1}\in \mathbf{R}^{L_i}$};

\draw[-Stealth, very thick] (3*\deltacol+3,0) -- (4*\deltacol-2.75,0);

%%%% PHI DEC
\node [rectangle, fill=white, minimum width=5em, minimum height=12em, draw, thick] at (4*\deltacol,0) {\LARGE$\Psi_\mathrm{\Large Align}$};
\draw[-Stealth, very thick] (4*\deltacol,-6) -- (4*\deltacol,-11) node [right, pos=0.70] {\Large$(\theta_i)_{i=1}^N$};

\draw[-, very thick] (0.68*\deltacol+3.75,-2.4) -- (1.9*\deltacol-4.3,-2.4);
\draw[-Stealth, very thick] (1.9*\deltacol-4.3,-2.5) -- (1.9*\deltacol-4.3,3.5*\deltachannelsy) -- (3.9*\deltacol-3,3.5*\deltachannelsy);
\draw[-Stealth, very thick] (1.9*\deltacol-4.3,-2.5) -- (1.9*\deltacol-4.3,0*\deltachannelsy) -- (2.05*\deltacol-3,0*\deltachannelsy);
\draw[-Stealth, very thick] (4.5*\deltacol,-13.75) -- (7.3*\deltacol-4.3,-13.75);

\draw[-, very thick] (4*\deltacol+2.75,0) -- (5*\deltacol-4.3,0);
\draw[-Stealth, very thick] (5*\deltacol-4.3,0) -- (5*\deltacol-4.3,1*\deltachannelsy) -- (5*\deltacol-3,1*\deltachannelsy);
\draw[-Stealth, very thick] (5*\deltacol-4.3,0) -- (5*\deltacol-4.3,-1*\deltachannelsy) -- (5*\deltacol-3,-1*\deltachannelsy);

%%%%% CHN 3

\node[inner sep=0pt] at (5*\deltacol,1*\deltachannelsy) {\includegraphics[width=8em]{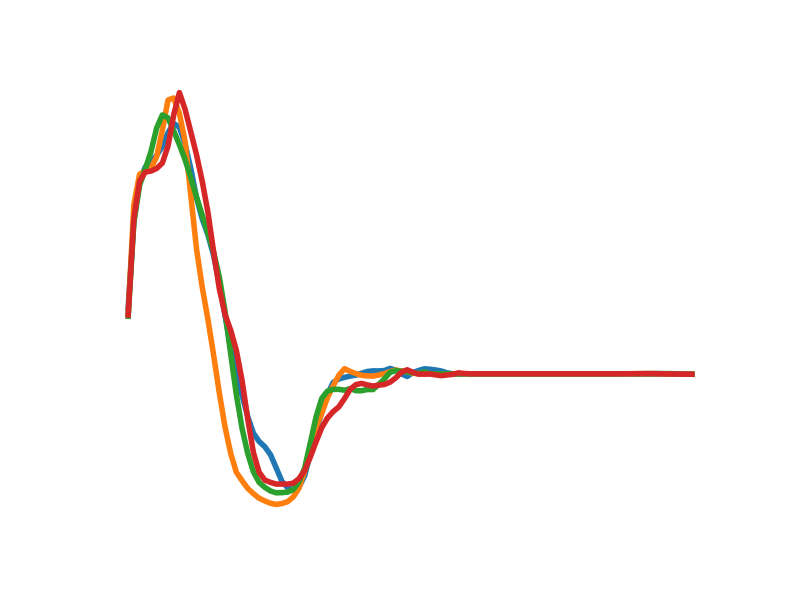}};

\node[inner sep=0pt] at (5*\deltacol,-1*\deltachannelsy) {\includegraphics[width=8em]{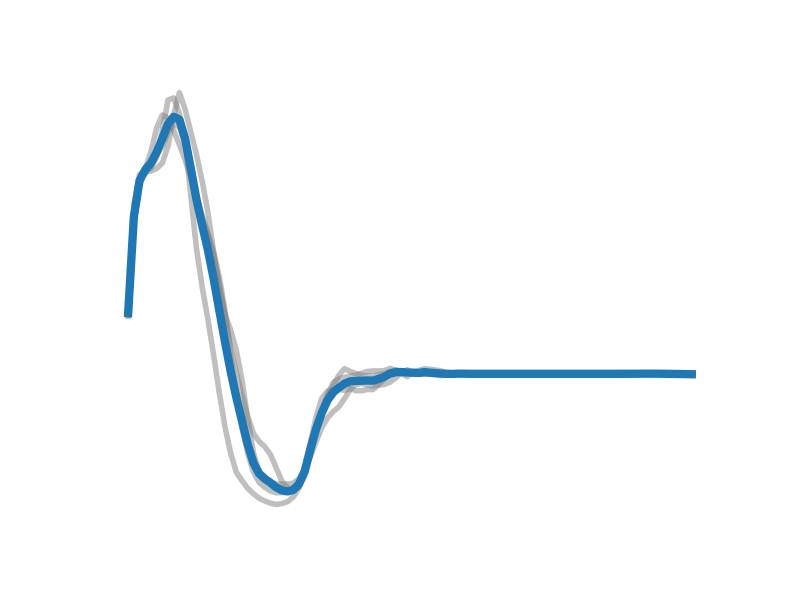}};

\draw[-, very thick] (5*\deltacol+3,1*\deltachannelsy) -- (5*\deltacol+4.3,1*\deltachannelsy) -- (5*\deltacol+4.3,0);
\draw[-Stealth, very thick] (5*\deltacol+4.3,0) -- (6*\deltacol-2.75,0);

\node[above,font=\small] (top3) at (5*\deltacol,1.5*\deltachannelsy) {\large Joint Alignment};
\node[above,font=\small] (top3) at (3*\deltacol,1.5*\deltachannelsy) {\large Dimensionality Reduction};

%%%%% PHI DEC

\node [trapezium, fill=white, trapezium stretches=true, trapezium angle=80, minimum width=12em, minimum height=5em, draw, thick, shape border rotate=90] (phidec) at (6*\deltacol,0) {\LARGE$\Psi_\mathrm{dec}$};

\def\deltachannelsy{4em}
\node[inner sep=0pt] (achn1) at (7.55*\deltacol,3*\deltachannelsy) {\includegraphics[width=10em]{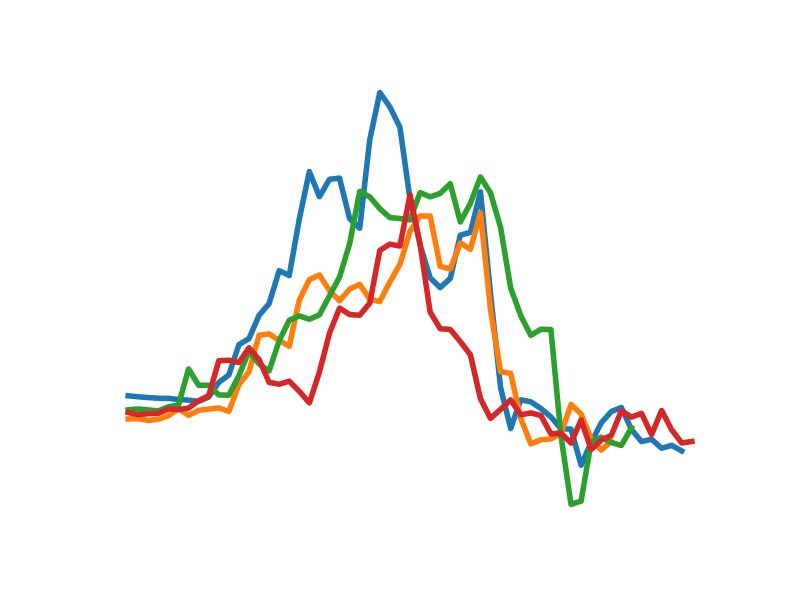}};

\node[inner sep=0pt] (achn2) at (7.55*\deltacol,1.8*\deltachannelsy) {\includegraphics[width=10em]{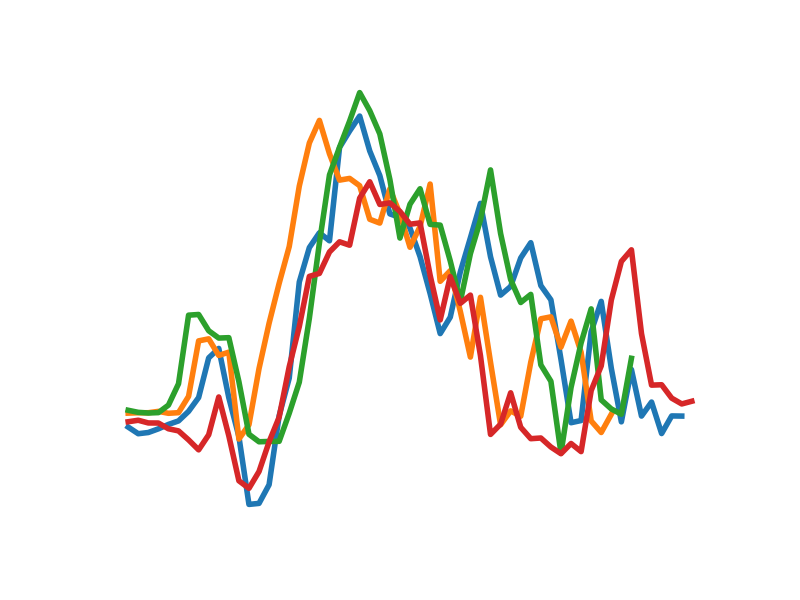}};

\path (7.55*\deltacol,0.5*\deltachannelsy) -- (7.55*\deltacol,-0.5*\deltachannelsy) node [font=\huge, midway, sloped] {$\dots$};

\node[inner sep=0pt] (achn3) at (7.55*\deltacol,-1.5*\deltachannelsy) {\includegraphics[width=10em]{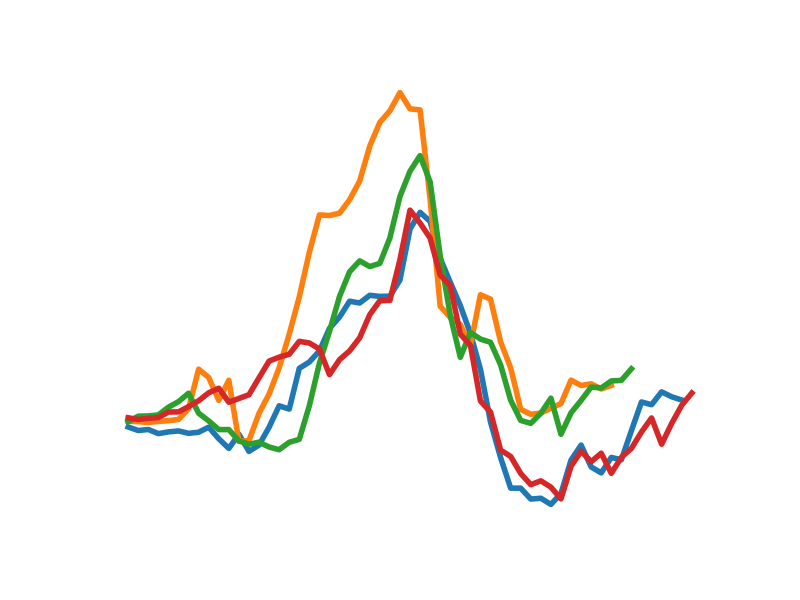}};

\node[above,font=\small] at (7.45*\deltacol,3.7*\deltachannelsy) {\large$(U_i\circ T^{\btheta_i})^{N}_{i=1}\in \mathbf{R}^{C \times L_i }$};

\node  (CPAB) [rectangle, fill=white, minimum width=5em, minimum height=5em, draw, thick] at (4*\deltacol,-14) {\large CPAB WARP};
%%%%%%%%%%%%%%%%%

\node (u_gal) [font=\large] at (6.55*\deltacol,-1.2) {\large$(\tilde{U}_i)_{i=1}^N$};

\node[font=\large] at (2*\deltacol,17) {\Large$\mathcal{L}_\mathrm{rec}((U_i)_{i=1}^N, ((\tilde{U}_i)_{i=1}^N, ((\theta_i)_{i=1}^N)$};

\node[font=\large] at (4.75*\deltacol,17) {\Large$\mathcal{L}_\mathrm{ICAE}(((Z_i)_{i=1}^N, \hat{Z}, ((\theta_i)_{i=1}^N))$};

\draw[-Stealth, thick, ronblue] (0.73*\deltacol,8.5) -- (0.73*\deltacol,14) -- (1.35*\deltacol,14) -- (1.35*\deltacol,15.5);

\draw[-Stealth, thick, ronblue] (6.31*\deltacol,0)--(6.59*\deltacol,0) -- (6.59*\deltacol,14) -- (2.25*\deltacol,14) -- (2.25*\deltacol,15.5);
\draw[-Stealth, thick, ronorange] (2.9*\deltacol,8.5) -- (2.9*\deltacol,13) -- (4.2*\deltacol,13) -- (4.2*\deltacol,15.5);

\draw[-, thick] (4*\deltacol,6) -- (4*\deltacol,12) node [right, pos=0.55] {\Large$(\theta_i)_{i=1}^N$};
\draw[-Stealth, thick, ronblue] (4*\deltacol,12) -- (2.8*\deltacol,12) -- (2.8*\deltacol,15.5);
\draw[-Stealth, thick, ronorange] (4*\deltacol,12) -- (5.55*\deltacol,12) -- (5.55*\deltacol,15.5);

\draw[-Stealth, thick, ronorange] (4.83*\deltacol,8.5) -- (4.83*\deltacol,15.5);

\begin{pgfonlayer}{background} 
    \node[fill=ronblue!10,dashed,very thick,draw=black,inner ysep=1.2em,inner xsep=0.8em, rounded corners=2mm] (bigbg) [fit = (CPAB) (top1) (achn3)] {};

    \node[fill=rongreen!10,dashed,very thick,draw=black,rounded corners=7mm,inner ysep=3em,inner xsep=0.7em]  (smallbg) [fit = (u_gal) (phienc) (phidec)] {};
    \node[above,font=\small] (top2) at (4.95*\deltacol,-0.525*\deltachannelsy) {\large$\hat{Z}\in  \mathbf{R}^{L}$} ;

\end{pgfonlayer}

\end{tikzpicture}
}
\caption{\textbf{Diffeomorphic Multitasking Autoencoder} (D-MTAE) for Temporal Prototype Learning, consists of: 1) $\Psi_{\text{enc}}$, an encoder for dimensionality reduction; 
2) $\Psi_{\text{Align}}$~\cite{Shapira:NIPS:2019:DTAN}, for joint alignment; and
3) $\Psi_{\text{Dec}}$, a decoder. The losses for JA and DR are $\mathcal{L}_{ICAE}$ and $\mathcal{L}_{rec}$ respectively. The feature extractor, $\phi$, could either by trained per dataset (\eg, CARL~\cite{Chen:CVPR:2022:CARL}) or a pretrained foundation model (\eg, DINO~\cite{Caron:ICCV:2021:dino}). 
}

\label{fig:arch1}
\end{figure*}

%% file: paper/tables/tab_notations.tex
\begin{table}[t]
\centering
\footnotesize
\caption{Table of notations.}
\begin{tabular}{@{}ll@{}}
\toprule
Symbol & Description \\ 
\midrule
$S_i$               & $i$-th video. \\
$s_t^i$             & Frame $t$ of video $S_i$. \\
$u_t^i=\phi(s_t^i)$ & Per-frame embedding. \\
$U_i=\{u_t^i\}_{t=1}^L$ & Embedded feature sequence, $\in\mathbb{R}^{C\times L}$. \\
$\widetilde{U}_i$ & Reconstructed embedded sequence. \\
$\btheta_i\in\mathbb{R}^d$  & Predicted warp parameters for $U_i$. \\
$T^{\btheta_i}$     & Parametric time-warp associated with $\btheta_i$. \\
$Z_i$               & Univariate representation. \\
$\widehat U / \widehat Z$        & Average aligned sequence. \\
\bottomrule
\end{tabular}
\label{tab:notations}
\end{table}

%% file: paper/fig_joint_alignment.tex
\begin{figure}[t]
  \centering
  \begin{subfigure}{0.39\textwidth}
  % [left, down, right, up]
    \includegraphics[trim = 5mm 0mm 5mm 12mm, clip, width=\textwidth]{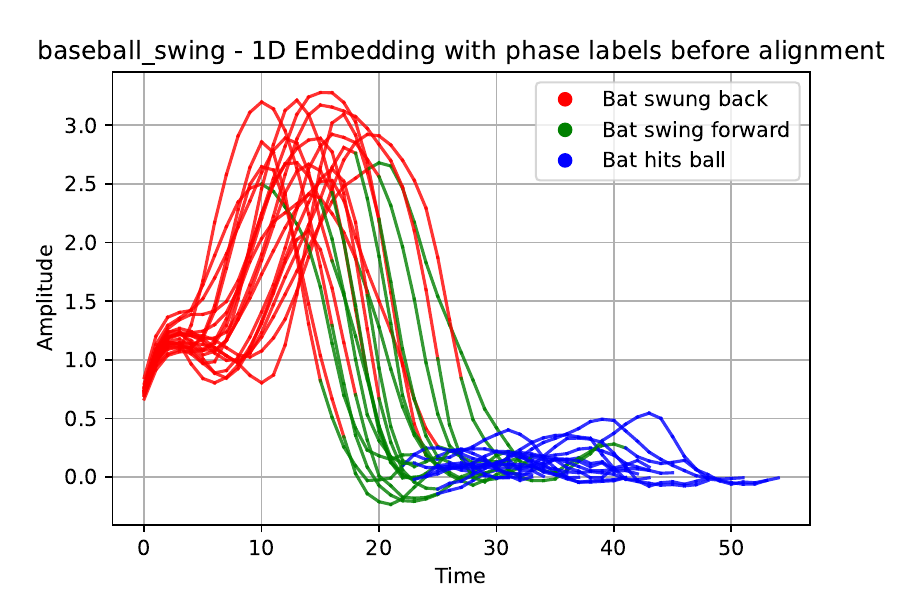}
    \caption{\emph{Baseball swing} learned 1D Latent representation.}
  \end{subfigure}
  \begin{subfigure}{0.39\textwidth}
    \includegraphics[trim = 5mm 0mm 5mm 12mm, clip, width=\textwidth]{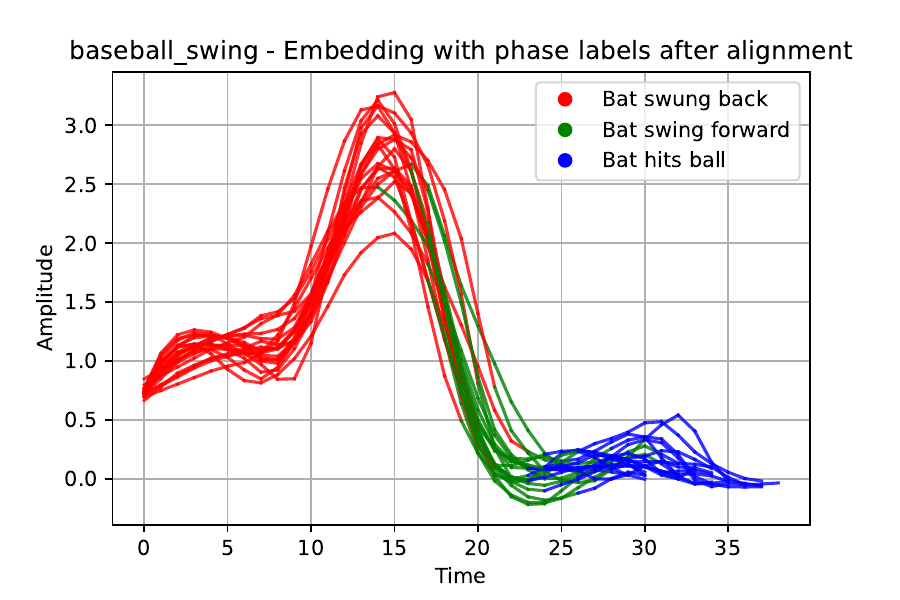}
    \caption{After synchronization.}
  \end{subfigure}

\caption{Univariate representations learned by TPL for 20 videos depicting a \emph{Baseball swing} colored by the phase labels, before (top) and after Synchronization (bottom).}\label{fig:JA:1D}
\end{figure}

%% file: paper/fig_genai.tex
\newcommand{\boxsize}{1.22}
\newcommand{\squarethick}{1.5pt}
\definecolor{green2}{RGB}{0,145,0}

\begin{figure*}[ht]
    \centering

    % ------------------------------
    % Row 1, Left Image
    % ------------------------------
    \begin{tikzpicture}
        % Place the image
        \node[anchor=south west, inner sep=0] (imgA) at (0,0)
        {\includegraphics[trim=1mm 1mm 1mm 1mm, clip, width=0.49\linewidth]{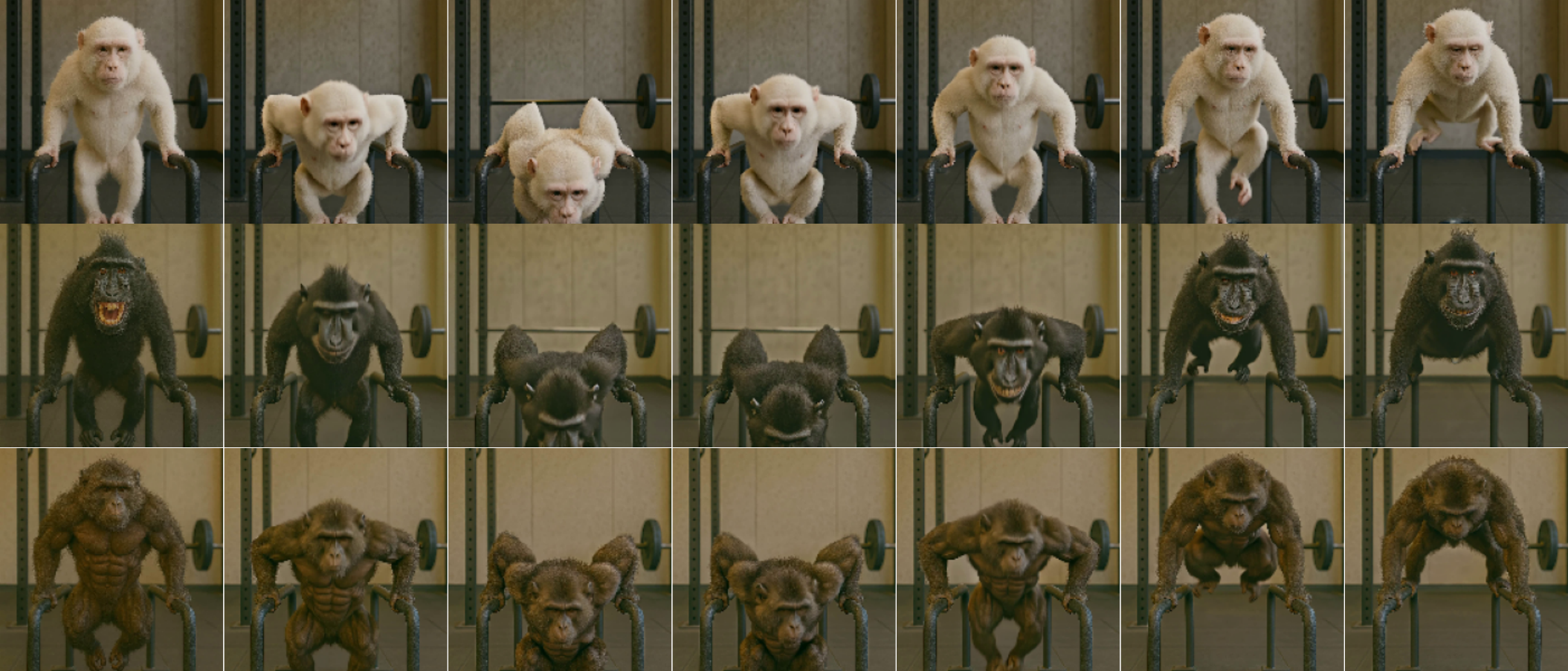}};
        % Draw one or more squares of side \boxsize. 
        % Change (1,1) to the coordinate where you want to place the bottom-left corner of the square.
        \draw[red, line width=\squarethick] (2.46,2.45) rectangle ++(\boxsize,\boxsize);
        \draw[red, line width=\squarethick] (3.68,2.45-\boxsize) rectangle ++(\boxsize,\boxsize);
        \draw[red, line width=\squarethick] (3.68,1.23-\boxsize) rectangle ++(\boxsize,\boxsize);
    \end{tikzpicture}
    %
    % ------------------------------
    % Row 1, Right Image
    % ------------------------------
    \begin{tikzpicture}
        % Place the image
        \node[anchor=south west, inner sep=0] (imgB) at (0,0)
        {\includegraphics[trim=1mm 1mm 1mm 1mm, clip, width=0.49\linewidth]{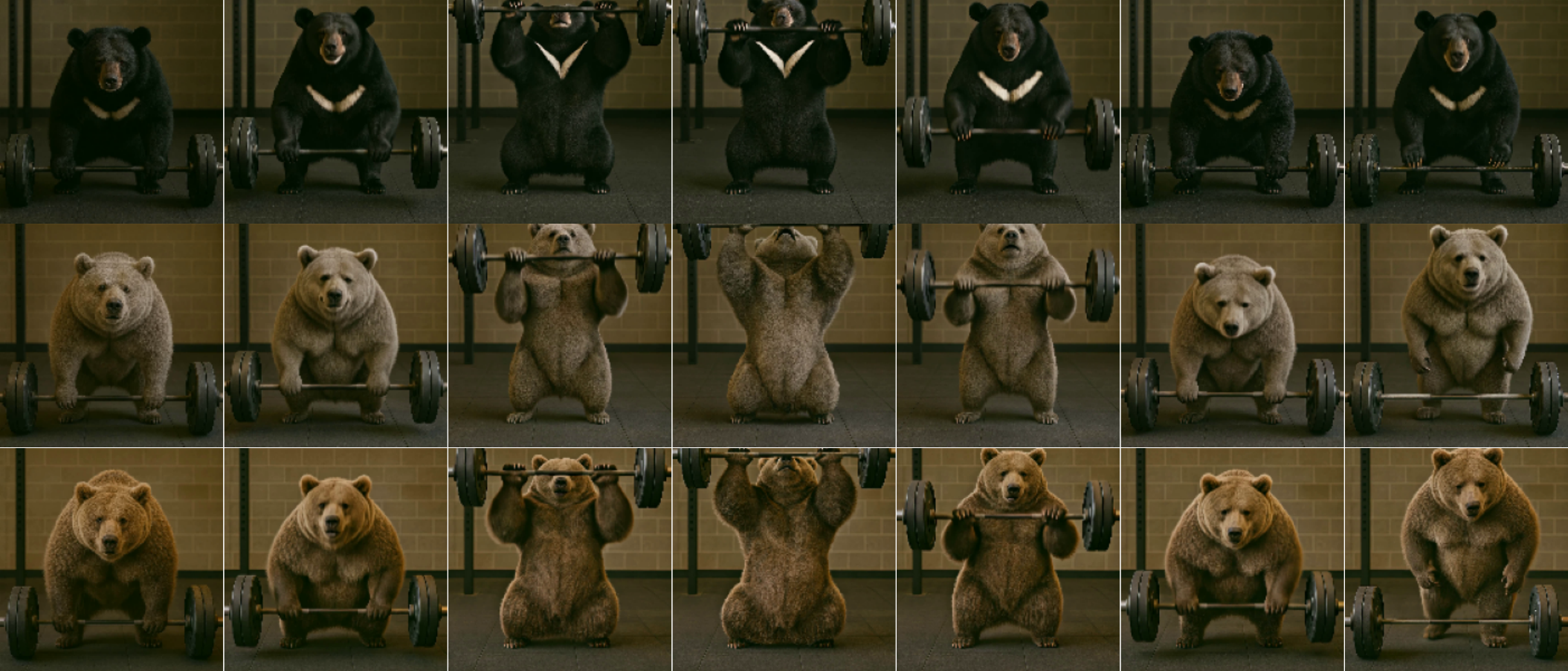}};
        % Overlay squares of side \boxsize
        \draw[red, line width=\squarethick] (2.46,2.45) rectangle ++(\boxsize,\boxsize);
        \draw[red, line width=\squarethick] (3.68,2.45-\boxsize) rectangle ++(\boxsize,\boxsize);
        \draw[red, line width=\squarethick] (3.68,1.23-\boxsize) rectangle ++(\boxsize,\boxsize);
    \end{tikzpicture}
    
    % ------------------------------
    % Row 2, Left Image
    % ------------------------------
  %  \raisebox{-0.00001cm}{
    \begin{tikzpicture}

            \node[anchor=south west, inner sep=0] (imgC) at (0,0)
        {\includegraphics[trim=1mm 1mm 1mm 1mm, clip, width=0.49\linewidth]{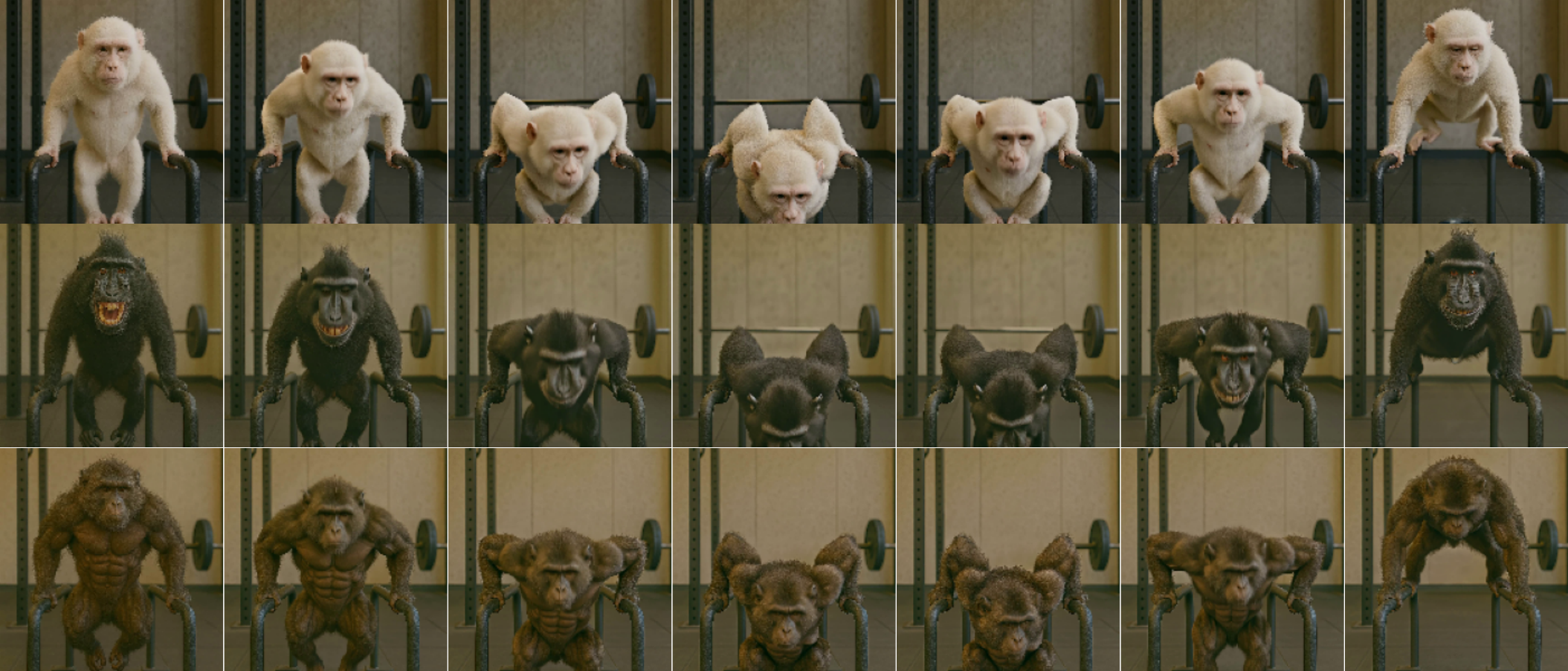}};
        
        % Overlay squares
        \draw[green2, line width=\squarethick] (3.7,2.45) rectangle ++(\boxsize,\boxsize);
        \draw[green2, line width=\squarethick] (3.7,2.45-\boxsize) rectangle ++(\boxsize,\boxsize);
        \draw[green2, line width=\squarethick] (3.7,1.23-\boxsize) rectangle ++(\boxsize,\boxsize);
    \end{tikzpicture}
 %   }%
    % ------------------------------
    % Row 2, Right Image
    % ------------------------------
    \begin{tikzpicture}
        \node[anchor=south west, inner sep=0] (imgD) at (0,0)
        {\includegraphics[trim=1mm 1mm 1mm 1mm, clip, width=0.49\linewidth]{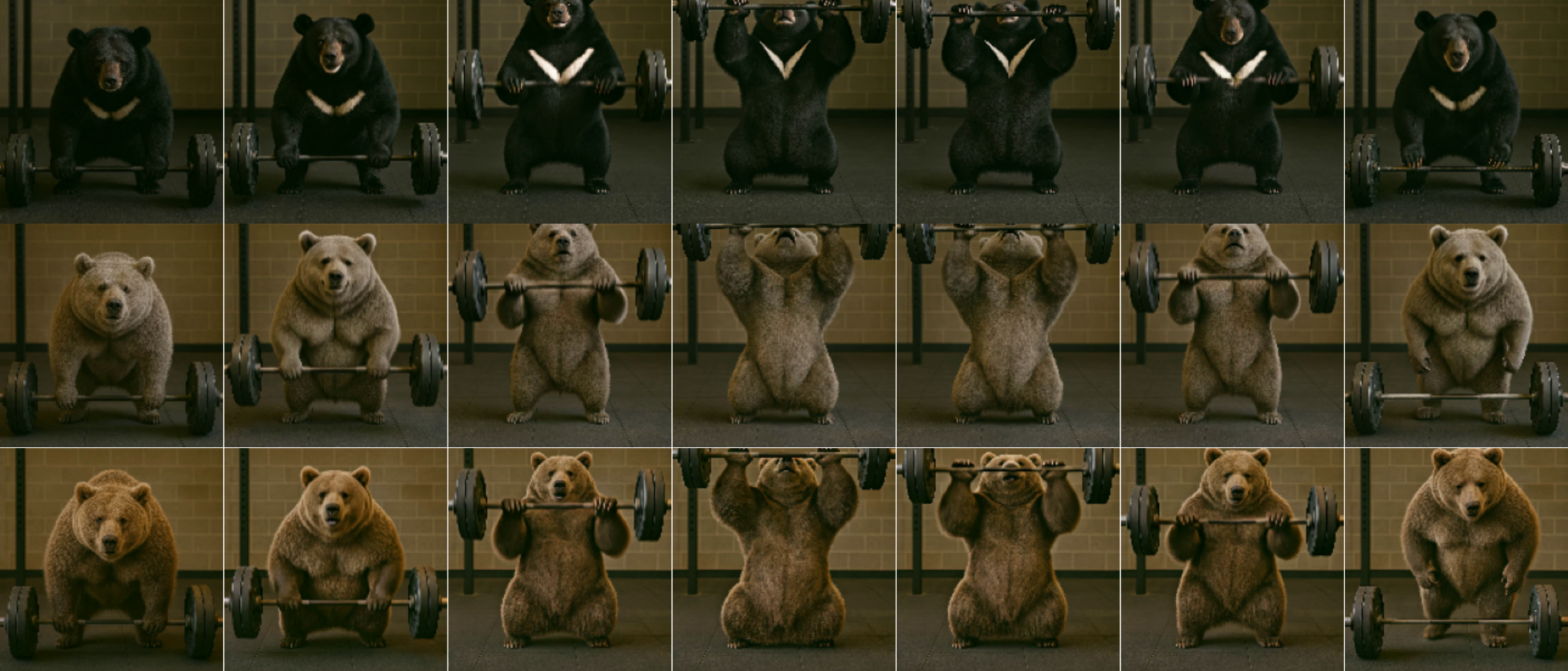}};
        
        % Overlay squares
        % Overlay squares of side \boxsize
        \draw[green2, line width=\squarethick] (3.7,2.45) rectangle ++(\boxsize,\boxsize);
        \draw[green2, line width=\squarethick] (3.7,2.45-\boxsize) rectangle ++(\boxsize,\boxsize);
        \draw[green2, line width=\squarethick] (3.7,1.23-\boxsize) rectangle ++(\boxsize,\boxsize);
    \end{tikzpicture}
    
    \caption{Examples from our GenerativeAI Multiple Video Synchronization (GenAI-MVS) dataset, showing seven equally spaced frames before (top) and after (bottom) synchronization. The first video (left) depicts a "monkey doing dips," and the second video (right) shows a "bear performing a deadlift." We highlight mismatches in the original videos in red, and TPL matching in green.  In both cases, alignment via TPL successfully synchronizes the key phases of the action progression.}
    \label{fig:genai:videos}
\end{figure*}
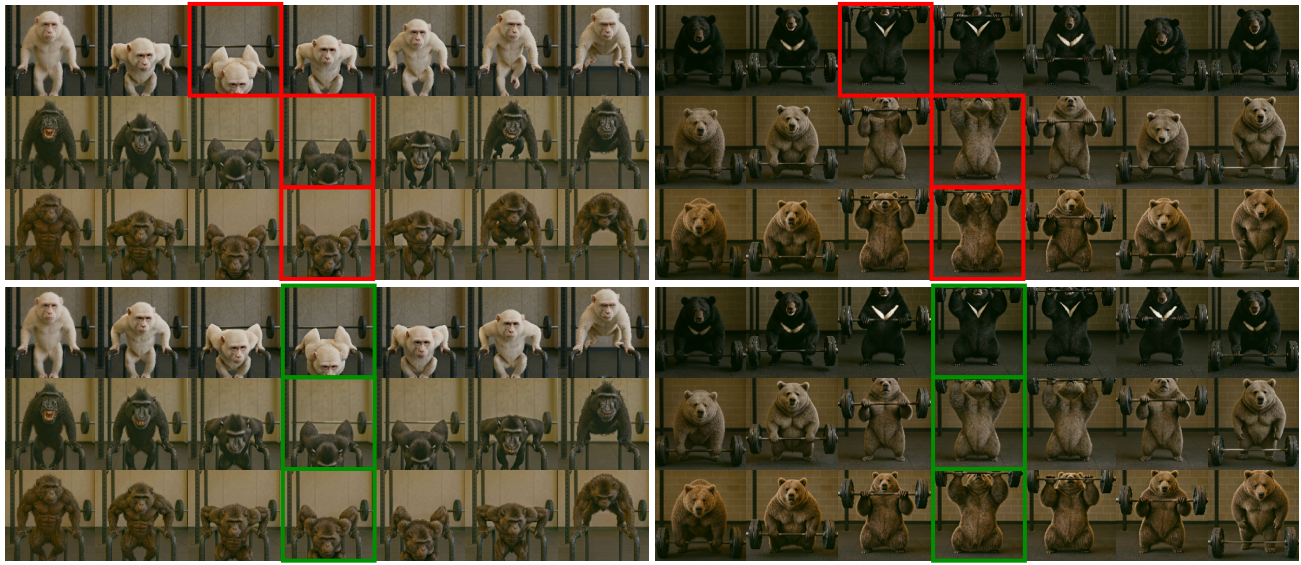

%% file: paper/results.tex
In this section, we present a series of experiments designed to demonstrate the effectiveness of TPL for multiple video synchronization (MVS). 
\subsection{Datasets}\label{Sub:Sec:Results:Baselines}
We evaluate TPL on the following datasets:
\begin{enumerate}
    \item \textbf{Pouring}~\cite{DBLP:journals/corr/SermanetLHL17}:
    A standard benchmark consisting of 84 videos of people pouring liquids into glasses. 
    
    \item \textbf{Penn Action}~\cite{zhang:ICCV:2013:penn}:
    This dataset contains 2326 videos of 15 different actions performed in the wild, varying in camera angles, lighting, action duration, backgrounds, and subjects, with phase-level annotations produced by~\cite{Debidatta:CVPR:2019:TCCL}.
    
    \item \textbf{Internet Video Dataset}~\cite{Dong:ECCV:2020:NetVideos}:
    A smaller dataset, similar to Penn Action, comprising 124 videos of 20 actions. We annotate the phases in the same manner as~\cite{Debidatta:CVPR:2019:TCCL}.

\item \textbf{GenAI Multiple Video Synchronization Dataset}:
We introduce a first-of-its-kind collection of AI-generated videos using \texttt{KlingAI} for the task of MVS (GenAI-MVS). For each action, a text prompt is composed, and an initial image is generated using \texttt{ChatGPT}. The image and prompt are then used as input to \texttt{Kling AI} to generate a video of the action. Multiple videos of the same action are generated in this manner, resulting in natural variation in both visual appearance and temporal execution. The dataset contains 5 classes and 82 hand-picked videos curated for MVS, each accompanied by phase progression annotations (see \textbf{SupMat} for more details).

\end{enumerate}

 %%%%%%%%%
% TABLE %
%%%%%%%%%
\subimport{./tables}{tab_sync_error}
\subsection{Evaluation Metrics}\label{sec:results:subsec:evaluation}
As stated in~\cite{Dave:ECCV:2024:sync}, existing benchmarks often rely on proxy tasks such as phase classification by a linear classifier or Kendall’s Tau for phase progression~\cite{Debidatta:CVPR:2019:TCCL}. These metrics had been shown to be affected by spurious correlations between the positional encoding of the model {(\eg, CARL~\cite{Chen:CVPR:2022:CARL})} and the phase labels. 
To evaluate alignment directly, we follow~\cite{Dave:ECCV:2024:sync} and introduce two metrics: \emph{Cycle-Back Consistency (CBC)}, which measures how well phase labels are preserved when warping videos to the prototype and then un-warping them back, and \emph{Phase Label Propagation (PLP)}, which measures alignment quality by transferring phase labels from a train-set prototype to test videos.  We still report phase classification and Kendall’s Tau for completeness, but CBC and PLP offer a clearer measure of real-world alignment performance:
\begin{itemize}[leftmargin=*]
    \item \textbf{Cycle-Back Consistency (CBC).} 
    Measures how well the prototype maintains phase information. The videos are warped to the prototype and label it according to their annotation. The prototype is then unwarped back to each video, and the phase labels are compared. A higher CBC indicates more accurate and robust synchronization.

    \item \textbf{Phase Label Propagation (PLP).} 
    Assesses alignment by transferring phase labels from a prototype to each test video. The better the alignment, the more accurately these labels will map onto the correct frames in the test video. PLP thus serves as a direct measure of alignment quality.

    \item \textbf{Phase Classification Accuracy.} 
    Assess the embedding quality by training a linear classifier on the per-frame embedding to predict the phase labels. 
    \item \textbf{Kendall’s Tau.} 
    A rank-correlation metric that evaluates the chronological order of phases across videos.
\end{itemize}

\subsection{Comparison with Multiple Sequence Alignment (MSA) Methods}
\label{subsec:compare_msa}
To evaluate our method's performance \wrt existing approaches, we align sets of videos depicting the same action using TPL and compare the results against 4 representative MSA methods:
Euclidean baseline (Euc.), where we zero-pad all videos to have the same length (according to the longest one) and compute metrics only on each video's valid regions.
DBA~\cite{Petitjean:2011:global}, SoftDBA~\cite{Cuturi:2017:soft}, and DTAN~\cite{Shapira:NIPS:2019:DTAN}. DBA and SoftDBA are optimization-based methods set to minimize the DTW and SoftDTW from the average sequence, respectively. For SoftDTW, we report the best results among $\gamma\in[0.01, 0.1, 1]$. DTAN is a learning-based method that predicts CPAB~\cite{Freifeld:ICCV:2015:CPAB} warps to minimize the JA loss. We evaluate DTAN with three losses: Within-Class Sum of Squares (WCSS), WCSS + Regularization (WCSS+Reg.), and the current state-of-the-art in time series averaging, DTAN+ICAE~\cite{Shapira:ICML:2023:rfdtan}. All DTAN models were trained using the closed-form CPAB gradient~\cite{Martinez:ICML:2022:closed}. 
We evaluate frame-level embedding from three feature extractors: 
1) CARL~\cite{Chen:CVPR:2022:CARL}, a video transformer that requires per-dataset training, 2) DINO-ViT--v2~\cite{Oquab:2023:dinov2}, a pre-train image transformer where we use the per-frame \texttt{CLS} token as the embedding, and 3) OpenCLIP~\cite{Cherti:CVPR:2023:openclip}, an open-source, more recent variation of CLIP~\cite{Radford:ICML:2021:CLIP}. 
We note that the available video foundation models (\eg, VideoMAE~\cite{Tong:NeurIPS:2022:videomae}) do not produce a per-frame embedding vector and were therefore excluded from this evaluation. 
We report CBC, LPL, and total runtime (training and inference time) on the Penn~\cite{zhang:ICCV:2013:penn}, Internet Videos~\cite{Dong:ECCV:2020:NetVideos}, and GenAI-MVS datasets. 

The results are presented in ~\autoref{tab:results}. We have found that internet videos~\cite{Dong:ECCV:2020:NetVideos} and GenAI-MVS did not have enough data to train CARL properly and are thus omitted.  We observe that TPL significantly outperforms all DTAN variants over all datasets and feature extractors. As discussed in~\autoref{subsec:method:preliminaries}, current DTAN formulations are ill-equipped to handle the real-world video embeddings.  TPL also outperforms DBA and SoftDBA across all benchmarks. While the margin in performance is less significant compared with DTAN, \textbf{TPL total runtime is $10$ times faster} than DBA and $4-5$ than SoftDBA on the largest dataset, Penn Action~\cite{zhang:ICCV:2013:penn}. GenAI-MSV results are further discussed in~\autoref{sec:method:subsec:genai}.

\subsection{Prototype-aligned Features}
\label{subsec:temporal_info}
To determine whether TPL prototypes capture meaningful phase progression, we evaluate phase classification accuracy and Kendall’s Tau rank correlation on the videos \emph{after} they have been aligned to the common prototype. By warping each video to the TPL prototype, we test whether these aligned representations provide more discriminative features for recognizing action phases. We conduct these experiments on both Penn Action~\cite{zhang:ICCV:2013:penn} and Pouring~\cite{Sermanet:ICRA:2018:pouring} datasets. We compare the prototyped-aligned features to standard benchmarks in video representation learning:  
TCC~\cite{Debidatta:CVPR:2019:TCCL},
GTA~\cite{Hadji:CVPR:2021:gta} ,
LAV~\cite{Haresh:CVPR:2021:LAV},
VAVA~\cite{Liu:CVPR:2022:VAVA},
VSP~\cite{zhang2023modeling}, and 
CARL\cite{Chen:CVPR:2022:CARL}. We report the results from their respective papers. 
The results, presented in~\autoref{tab:phase_classification}, indicates that TPL-aligned representations are on-par with VSP and CARL, the two strongest baselines. As mentioned in~\autoref{sec:results:subsec:evaluation}, these metrics are not ideal for assessing alignment quality. However, 
these findings indicate that the synchronized embeddings retain their temporal information after alignment. 
%%%%%%%%%
% TABLE %
%%%%%%%%%
\subimport{./tables}{tab_res1}

\subsection{Frame Retrieval Efficiency}
\label{subsec:frame_retrieval}
TPL’s MVS facilitates faster frame retrieval than
  standard KNN frame-retrieval approach, where each frame from each test video is compared to all frames in all videos in the train set. This implies $O(N_{\textrm{train}}N_{\textrm{test}}L^2)$ (assuming fixed-length $L$ for simplicity).  In contrast, video synchronization allows this process to be linear in $L$. This advantage arises because TPL establishes a single temporal reference for the action progression, allowing for direct frame lookup at each time step. For evaluation, we perform 1-NN frame retrieval on Penn using CARL's embedding and report the phase classification accuracy and runtime (including inference time for TPL).
As shown in~\autoref{tab:knn}, performing frame retrieval only between synchronized frames is \textbf{125 times faster} than a full KNN search ($0.24$ [sec]  and $30$ [sec], respectively).

%%%%%%%%%
% TABLE %
%%%%%%%%%
\subimport{./tables}{tab_knn_timing}

\subsection{Generalizing to Generative AI Videos}
\label{sec:method:subsec:genai}

Beyond real-world footage, we also study the effectiveness of TPL on synthetic videos generated via a combination of \texttt{ChatGPT} and \texttt{KlingAI}. For each action category, we compose a detailed text prompt and generate an initial reference image using \texttt{ChatGPT}. This image and prompt are then used as input to \texttt{KlingAI} to produce a video illustrating the target action. Finally, we annotate the phase progression in each video similarly to~\cite{Debidatta:CVPR:2019:TCCL}.
A key challenge in creating this dataset was identifying videos suitable for MVS. Current video generators often produce truncated action progressions, omit essential phases, or generate clips that do not depict the intended action. After filtering out problematic samples, we retained a diverse set of videos that still pose realistic alignment challenges.

An example of AI-generated video synchronization via TPL is shown in~\autoref{fig:genai:videos}. We display seven equally spaced frames taken from the original (top) and synchronized (bottom) sequences. For instance, in the ``\texttt{Bear deadlift}'' example, TPL successfully aligns key motion phases, including the moment the bear begins lifting and reaches the upright position, demonstrating improved temporal coherence.
We also report the CBC and PLP for the MSA methods and report them in ~\autoref{tab:results}.
TPL achieves higher CBC and Phase Label Propagation PLP than traditional Soft/DBA and DTAN.
These results highlight the ability of TPL to extend beyond conventional, human-recorded video sources to the emerging domain of AI-generated content, providing a robust solution for synchronizing multiple generative clips depicting the same action.

\subsection{Ablation Study}
\label{subsec:ablation}
\autoref{tab:ablation} shows the ablation study for Penn Action using the Euclidean distance as a baseline.
Only using the alignment network (without the 1D bottleneck) yields improvement in both CBC and PLP. 
However, introducing only the encoder diminishes the performance significantly, indicating the importance of reconstruction for stable training, as seen by the significant improvement in results when using the decoder.
Finally, enforcing a median-length prototype gives the full TPL framework that achieves the best overall results.

%%%%%%%%%
% TABLE %
%%%%%%%%%
\subimport{./tables}{tab_ablation}

%% file: paper/tables/tab_sync_error.tex
\begin{table*}[ht]
    \centering
    \scriptsize
    \caption{Comparison of different features and alignment methods on Penn Action, Internet Videos, and Gen AI. 
    We report the alignment objective to minimize (Obj.), Cycle-Back Consistency (CBC), Phase Label Propagation (PLP), 
    and total runtime (Time) in seconds.}
    \label{tab:results}
    \begin{tabular}{llc|ccc|ccc|ccc}
        \toprule
        & & & \multicolumn{3}{c|}{\textbf{Penn Action}} & \multicolumn{3}{c|}{\textbf{Internet Videos}} & \multicolumn{3}{c}{\textbf{GenAI-MVS}} \\
        \cmidrule(lr){4-6}\cmidrule(lr){7-9}\cmidrule(lr){10-12}
        \textbf{Features} & \textbf{Method} & \textbf{Obj.} 
        & \textbf{CBC} & \textbf{PLP} & \textbf{Time} 
        & \textbf{CBC} & \textbf{PLP} & \textbf{Time}
        & \textbf{CBC} & \textbf{PLP} & \textbf{Time} \\
        %\midrule
        \midrule
        \multirow{6}{*}{\shortstack{CARL \\ + dataset training}} 
        Baseline & Euc.     & Euc.        & 0.621 & 0.607 & 0.65 & N/A & N/A & N/A &  N/A & N/A & N/A \\

        & DTAN     & WCSS        & 0.42 & 0.415 & 588 & N/A & N/A & N/A & N/A & N/A & N/A \\
        & DTAN     & WCSS + Reg. & 0.647 & 0.625 & 710 & N/A & N/A & N/A & N/A & N/A & N/A \\
        & DTAN     & ICAE        & 0.773 & 0.765 & 579 & N/A & N/A & N/A & N/A & N/A & N/A \\
        & DBA      & DTW         & 0.947 & 0.925 & 2345 & N/A & N/A & N/A & N/A & N/A & N/A \\
        & SoftDTW  & SoftDTW     & 0.944 & 0.926 & 978 & N/A & N/A & N/A & N/A & N/A & N/A \\
        & TPL (ours) & $\mathcal{L}_{\textrm{TPL}}$ &
        \textbf{0.962} & \textbf{0.939} & 482 &N/A & N/A & N/A & N/A & N/A & N/A \\
        \midrule
        \multirow{6}{*}{\shortstack{DINO-ViT \\(`off-the-shelf')}} 
        & DTAN     & WCSS        & 0.418 & 0.419 & 599 & 0.712 & 0.63 & 215 &  0.759 & 0.768 & 105 \\
        & DTAN     & WCSS + Reg. & 0.591 & 0.604 & 707 & 0.764 & 0.647 & 224 & 0.762  & 0.752  & 110 \\
        & DTAN     & ICAE        & 0.572 & 0.578 & 639 & 0.874 & 0.683 & 232 & 0.833  & 0.740  & 114 \\
        & DBA      & DTW         & 0.756 & 0.773 & 5415 & 0.882 & 0.871 & 25 & 0.886 & 0.906 & 62 \\
        & SoftDTW  & SoftDTW     & 0.750 & 0.777 & 3010 & 0.883 & 0.872 & 53 & 0.890 & 0.908 & 58 \\
        & TPL (ours) & $\mathcal{L}_{\textrm{TPL}}$ 
        & \textbf{0.788} & \textbf{0.803} & 534 & \textbf{0.912} & \textbf{0.907} & 112 &
        \textbf{0.953} & \textbf{0.933} & 108 \\
        \midrule
        \multirow{6}{*}{\shortstack{OpenCLIP \\(`off-the-shelf')}} 
        & DTAN     & WCSS        & 0.418 & 0.419 & 629 & 0.792 & 0.7 & 223 & 0.834 & 0.803 & 104 \\
        & DTAN     & WCSS + Reg. & 0.636 & 0.615 & 736 & 0.83 & 0.742 & 229 & 0.754 & 0.748 & 112 \\
        & DTAN     & ICAE        & 0.704  & 0.688 & 680 & 0.858 & 0.742 & 237 & 0.535 & 0.614 & 110 \\
        & DBA      & DTW         & 0.807 & 0.808 & 4692 & 0.885 & 0.842 & 33 & 0.918 & 0.929 & 63 \\
        & SoftDBA  & SoftDTW     & 0.859 & 0.831& 1882 & 0.901 & 0.833 & 34 &  0.926&  0.921 & 68 \\
        & TPL (ours) & $\mathcal{L}_{\textrm{TPL}}$ & \textbf{0.873} & \textbf{0.857} & 587
        & \textbf{0.916} & \textbf{0.902} & 146 & \textbf{0.942} & \textbf{0.946} & 115 \\
        \bottomrule
    \end{tabular}
\end{table*}

%% file: paper/tables/tab_res1.tex
\begin{table}[ht]
\centering
\caption{Phase classification accuracy (Acc.) \& Kendall's Tau ($\tau$).
 Positional embedding is indicated (Pos. Emb.).}
\scriptsize
\begin{tabular}{@{} c l cc cc @{}}
\toprule
\textbf{Pos.\ Emb.} 
  & \textbf{Method} 
  & \multicolumn{2}{c}{\textbf{Penn Action}} 
  & \multicolumn{2}{c}{\textbf{Pouring}} 
\\
\cmidrule(lr){3-4}\cmidrule(lr){5-6}
& & Acc.\ & $\tau$ & Acc.\ & $\tau$ \\
\midrule
\xmark & TCC~\cite{Debidatta:CVPR:2019:TCCL} & 74.39 & 0.623 & 86.14 & 0.670 \\
\xmark & GTA~\cite{Hadji:CVPR:2021:gta}      & 78.90 & 0.654 & 85.16 & 0.750 \\
\xmark & LAV~\cite{Haresh:CVPR:2021:LAV}    & 78.68 & 0.805 & 92.84 & 0.856 \\
\xmark & VAVA~\cite{Liu:CVPR:2022:VAVA}     & 84.48 & 0.805 & 92.84 & 0.875 \\
\cmark & VSP~\cite{zhang2023modeling}      & 93.12 & 0.986 & 93.85 & 0.990 \\
\cmark & CARL~\cite{Chen:CVPR:2022:CARL}    & 93.07 & 0.985 & 93.73 & 0.992 \\
\cmark & TPL (ours)                         & \textbf{93.31} & \textbf{0.990} 
                                            & \textbf{93.88} & \textbf{0.993} \\
\bottomrule
\end{tabular}
\label{tab:phase_classification}
\end{table}

% \definecolor{mygray}{RGB}{90,90,90}

% \begin{table}[ht]
% \centering
% \caption{Phase classification accuracy (Acc.) \& Kendall's Tau ($\tau$).}
% \scriptsize
% \begin{tabular}{lccccc}
% \toprule
% \textbf{Method} & \multicolumn{2}{c}{\textbf{Penn Action}} & \multicolumn{2}{c}{\textbf{Pouring}} & \\
% \cmidrule(lr){2-3} \cmidrule(lr){4-5}
% & Acc. &  $\tau$ & Phase Classification &  $\tau$ & \textbf{} \\
% \midrule

% TCC~\cite{Debidatta:CVPR:2019:TCCL} & 74.39 & 0.623 &86.14 &0.67 & \\
% GTA~\cite{Hadji:CVPR:2021:gta} & 78.9 & 0.654  & 85.16 & 0.75 & \\
% LAV~\cite{Haresh:CVPR:2021:LAV} & 78.68 & 0.805 & 92.84 & 0.856& \\
% VAVA~\cite{Liu:CVPR:2022:VAVA} & 84.48 & 0.805 & 92.84 & 0.875  & \\

% VSP~\cite{zhang2023modeling}  &  93.12 & 0.986 & 93.85 & 0.990  & \\
% CARL~\cite{Chen:CVPR:2022:CARL} &  93.07 & 0.985 & 93.73 & 0.992  & \\

% TPL (ours)&  \textbf{93.31} & \textbf{0.990} & \textbf{93.88} &  \textbf{0.993}  & \\

% \bottomrule
% \end{tabular}\\
% \label{tab:phase_classification}
% \end{table}

%% file: paper/tables/tab_knn_timing.tex
\begin{table}[t]
    \centering
    \footnotesize
    \captionof{table}{Unsynchronized vs. synchronized  Nearest-neighbor frame retrieval comparison on Penn Action.}
    \begin{tabular}{lcccc}
        \toprule
        \textbf{Method} &  \textbf{Complexity} &  \textbf{Time (sec)}\\
        \midrule
        Unsynchronized        & $O(N_{\textrm{train}}N_{\textrm{test}} L^2)$ &  30.1  \\
        synchronized (Ours) & $O(N_{\textrm{train}}N_{\textrm{test}} L)$       &  \textbf{0.24}  \\
        \bottomrule
    \end{tabular}
    \label{tab:knn}
\end{table}

%\begin{table}[t]
%    \centering
%    \footnotesize
%    \captionof{table}{Un-synchronized vs. synchronized  Nearest-%neighbor frame retrieval comparison on Penn Action.}
%    \begin{tabular}{lcccc}
%        \toprule
%        \textbf{Method} &  \textbf{Complexity} & \textbf{Acc.} & %\textbf{Time (sec)}\\
%        \midrule
%        Un-synchronized        & %$O(N_{\textrm{train}}N_{\textrm{test}} L^2)$ & 92.9 & 30.1  \\
%        synchronized (Ours) & $O(N_{\textrm{train}}N_{\textrm{test}} %L)$       & \textbf{93.3} & \textbf{0.24}  \\
%        \bottomrule
%    \end{tabular}
%    \label{tab:knn}
%\end{table}

%% file: paper/tables/tab_ablation.tex
\begin{table}[t]
    \centering
    \footnotesize
     \captionof{table}[Ablation study evaluation] {Ablation study evaluation on Penn Action.}
    \label{tab:ablation}
    \begin{tabular}{l cc}
    \toprule
    \textbf{Condition} & \textbf{CBC} & \textbf{PLP}\\
    \midrule
        Baseline (Euclidean)       & 64.6\% & 63.5\% \\
        No 1D Bottleneck           & 80.4\% & 81.5\% \\
        Encoder Only               & 56.5\% & 51.3\% \\
        + Decoder, Standard ICAE   & 97.3\% & 96.7\%\\
    + ICAE with Median Length (TPL)  & 100\% & 100\%\\
        \bottomrule
    \end{tabular}
\end{table}

%% file: paper/conclusion.tex
We introduced \textbf{Temporal Prototype Learning (TPL)}, a novel framework for synchronizing multiple videos from different scenes without relying on a reference by simultaneously reducing high-dimensional embeddings to a univariate representation. TPL outperforms existing alignment methods on a range of real-world datasets, while also generalizing effectively to AI-generated content exhibiting diverse visual styles and timing variations. Moreover, its prototype-based alignment yields faster frame retrieval and requires fewer pairwise comparisons, making TPL well-suited for large-scale video analytics.

%% file: supmat/dataset.tex
\subsection*{Gen-MVS Dataset Details}
\label{sec:supmat:genmvs}

Gen-MVS is a dataset of 82 AI-generated videos synthesized using prompts and images via \texttt{ChatGPT} and \texttt{KlingAI}, as described in the main paper. It contains 5 action classes with natural variation in visual style, motion speed, and subject identity (e.g., different instances of the same animal category, such as a bulldog and a German shepherd performing the same action). Each video is annotated with a \textit{Start}, \textit{End}, and one class-specific key event, supporting evaluation of multi-video synchronization (MVS), as summarized in \autoref{tab:gen-mvs-details}.

\begin{table*}[!h]
\setlength{\tabcolsep}{0.3em}
\centering
\caption{List of all key events in the Gen-MVS dataset.
Each action has a \textit{Start} event and \textit{End} event in addition to the key event.}
\footnotesize{
    \begin{tabular}{l|c|p{7.3cm}|c|c}
        \textbf{Action} & \textbf{\#phases} & \textbf{Key Event} & \textbf{Train set} & \textbf{Val set} \\
        \midrule
        Bench-press & 2 & Bar fully down     & 9  & 5 \\
        Deadlift    & 2 & Bar fully lifted   & 11 & 6 \\
        Dips        & 2 & Elbows at 90°      & 12 & 6 \\
        Pullups     & 2 & Chin above bar     & 11 & 5 \\
        Pushups     & 2 & Head at floor      & 11 & 6 \\
        \bottomrule
    \end{tabular}
}
\label{tab:gen-mvs-details}
\vspace{-1em}
\end{table*}

\paragraph{Annotation.}
All videos were manually filtered for visual and temporal quality, and annotated with per-video phase progression and key event frames. These annotations are used for both supervision and alignment evaluation.
\newpage

%% file: supmat/results.tex
\subsection*{Joint alignment of video embeddings}
The Temporal Prototype Learning (TPL) framework seeks to learn the joint alignment and temporal prototypes of video action sequences. It is achieved via the Diffeomorphic Multitasking Autoencoder (D-MTAE). The output of the encoder, $\Psi{_\text{encoder}}$, is a 1-D representation of the multi-channel inputs, which are then jointly-aligned by a Diffeomorphic Temporal Alignment Net (DTAN)~\cite{Shapira:NIPS:2019:DTAN}, $\Psi{_\text{DTAN}}$. ~\autoref{fig:JA:1} presents the alignment results on the \emph{Baseball swing} and \emph{Golf swing} actions.

\subimport{./}{fig_joint_alignment}
\subsection*{Multiple video synchronization}
We provide qualitative results demonstrating the effectiveness of our Temporal Prototype Learning (TPL) for synchronizing multiple unsynchronized videos of the same action. TPL leverages learned temporal embeddings to align sequences with high temporal fidelity, even under challenging conditions such as viewpoint variation and subtle temporal misalignments. As shown in \autoref{fig:penn_action:squat}, \autoref{fig:penn_action:baseball_pitch}, \autoref{fig:penn_action:baseball_swing}, and \autoref{fig:penn_action:tennis_forehand}, TPL accurately estimates temporal offsets and brings semantically corresponding frames into alignment across multiple video sources. 
\subimport{./}{sync}

%% file: supmat/sync.tex
\newcommand{\boxwidth}{3.4}   % rectangle width
\newcommand{\boxheight}{2.6}  % rectangle height
\definecolor{green2}{RGB}{0,145,0}

\newcommand{\boxsizepenn}{2.5}
\begin{figure}[H]
    \centering

    % ------------------------------
    % Squat Original (Unsynced)
    % ------------------------------
    
    \begin{tikzpicture}
    \node[anchor=south west, inner sep=0] (imgA) at (0,0)
    {\includegraphics[trim=0mm 0mm 0mm 0mm,  clip, width=0.98\linewidth]{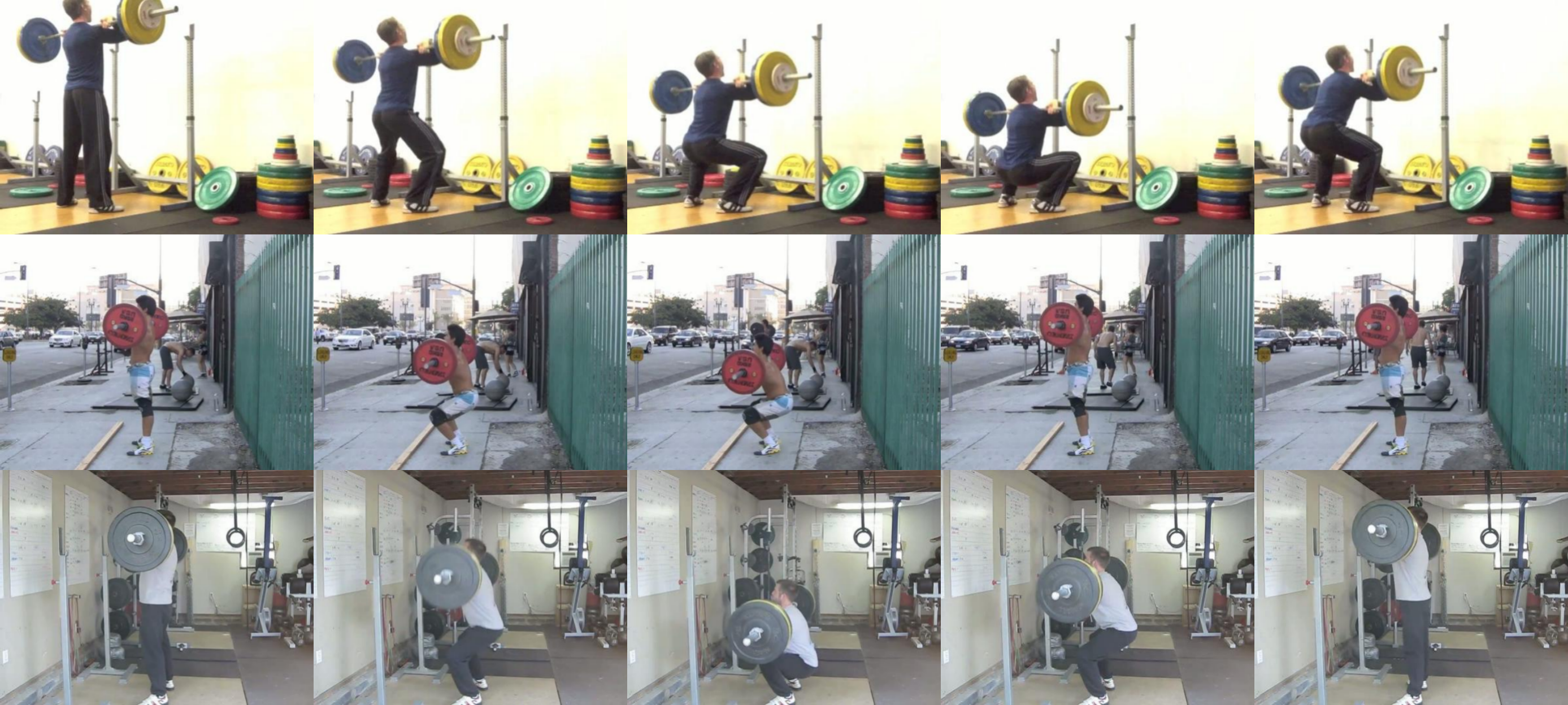}};
    
        \draw[red, line width=\squarethick] (10.2,5.2) rectangle ++(\boxwidth,\boxheight);

        % Overlay squares

        \draw[red, line width=\squarethick] (6.8,5.2-\boxheight) rectangle ++(\boxwidth,\boxheight);
        \draw[red, line width=\squarethick] (6.8+\boxwidth,0) rectangle ++(\boxwidth,\boxheight);
    \end{tikzpicture}
    
\hspace{2cm}
        
    % ------------------------------
    % Bear Synced
    % ------------------------------
    \begin{tikzpicture}
    \node[anchor=south west, inner sep=0] (imgA) at (0,0)
    {\includegraphics[trim=0mm 0mm 0mm 0mm, clip, width=0.98\linewidth]{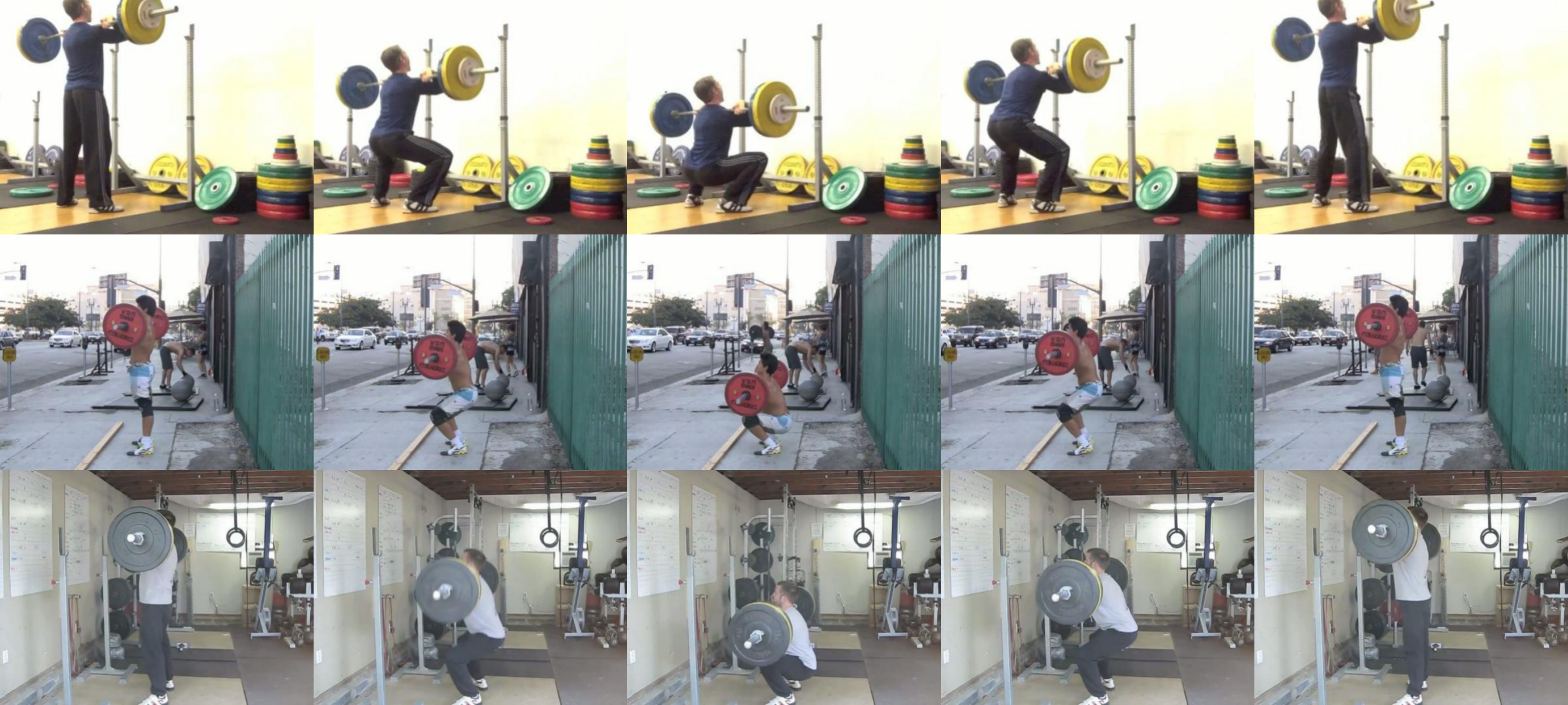}};
        
        % Overlay squares
        \draw[green2, line width=\squarethick] (6.8,5.2) rectangle ++(\boxwidth,\boxheight);
        \draw[green2, line width=\squarethick] (6.8,5.2-\boxheight) rectangle ++(\boxwidth,\boxheight);
        \draw[green2, line width=\squarethick] (6.8,0) rectangle ++(\boxwidth,\boxheight);
    \end{tikzpicture}
       \captionsetup{type=figure}

        \captionof{figure}[TPL Alignment: Squat]{Examples from  Penn Action dataset (squat action), showing five equally spaced frames before and after synchronization. We highlight mismatches in the original videos in red, and TPL matching in green. Alignment via TPL successfully synchronizes the key phases of the squat action.}
    \label{fig:penn_action:squat}
\end{figure}

\begin{figure}[H]
    \centering

    % ------------------------------
    % Squat Original (Unsynced)
    % ------------------------------
    \begin{tikzpicture}
    \node[anchor=south west, inner sep=0] (imgA) at (0,0)
    {\includegraphics[trim=0mm 0mm 0mm 0mm,  clip, width=0.98\linewidth]{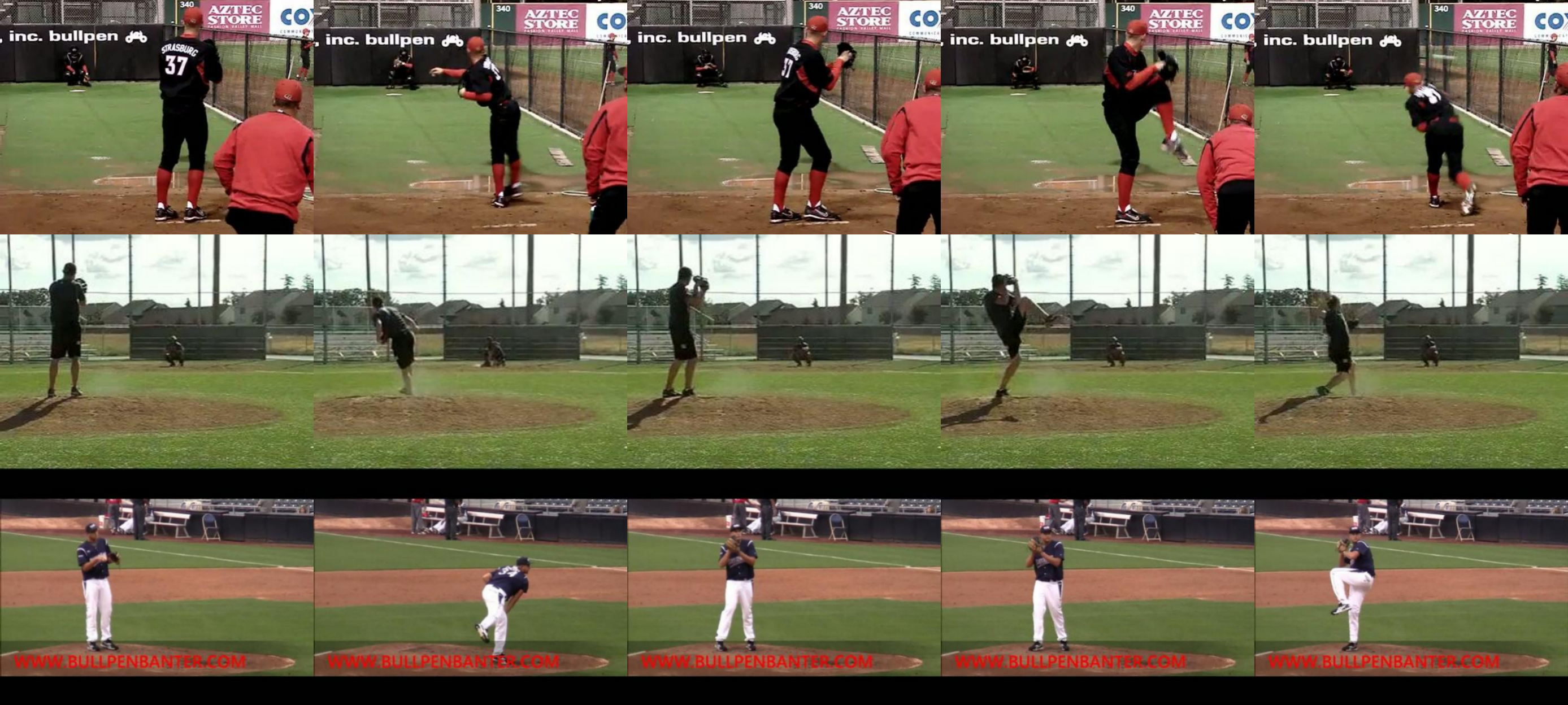}};
    
        \draw[red, line width=\squarethick] (10.2,5.2) rectangle ++(\boxwidth,\boxheight);

        % Overlay squares

        \draw[red, line width=\squarethick] (10.2,5.2-\boxheight) rectangle ++(\boxwidth,\boxheight);
        \draw[red, line width=\squarethick] (10.2+\boxwidth,0) rectangle ++(\boxwidth,\boxheight);
    \end{tikzpicture}
    
\hspace{2cm}
        
    % ------------------------------
    % Bear Synced
    % ------------------------------
    \begin{tikzpicture}
    \node[anchor=south west, inner sep=0] (imgA) at (0,0)
    {\includegraphics[trim=0mm 0mm 0mm 0mm, clip, width=0.98\linewidth]{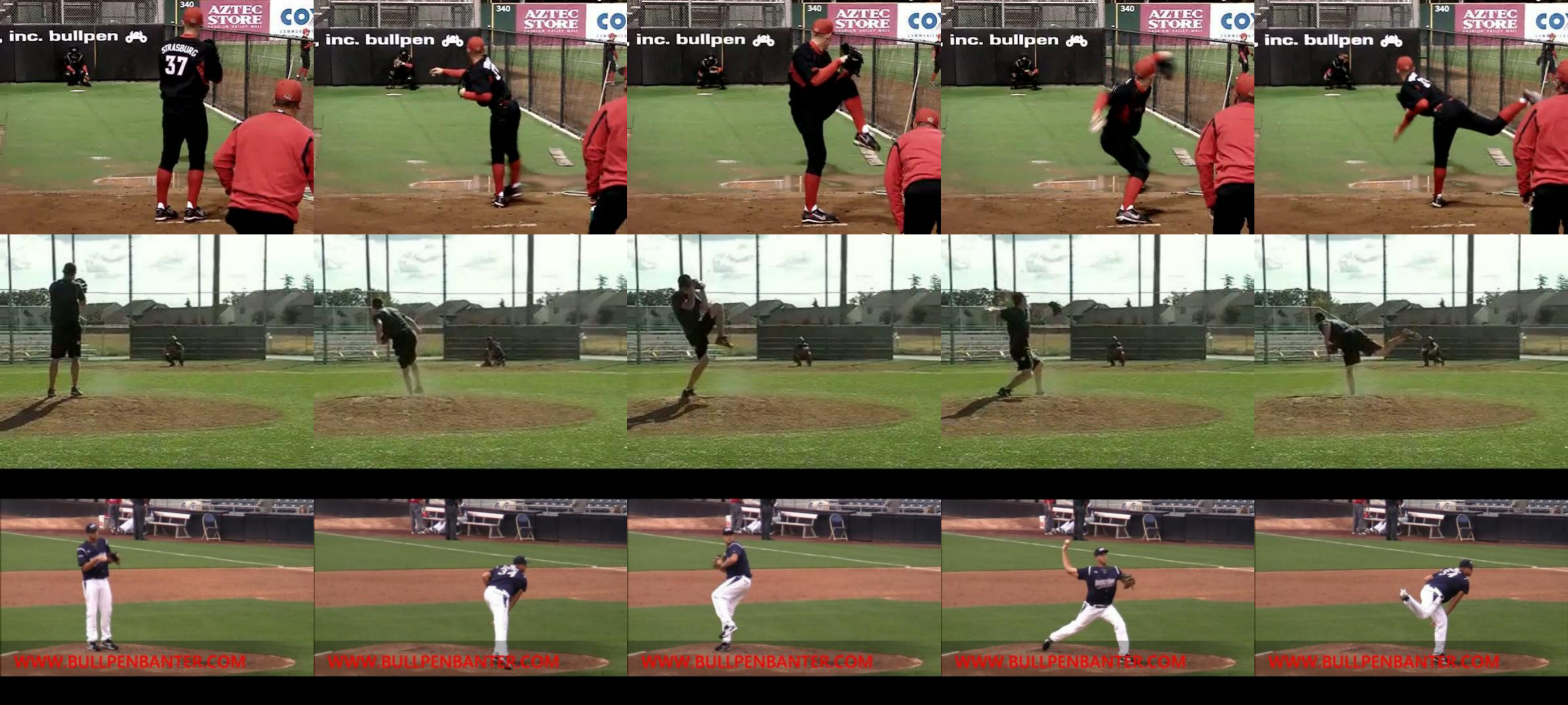}};
        
        % Overlay squares
        \draw[green2, line width=\squarethick] (6.8,5.2) rectangle ++(\boxwidth,\boxheight);
        \draw[green2, line width=\squarethick] (6.8,5.2-\boxheight) rectangle ++(\boxwidth,\boxheight);
        \draw[green2, line width=\squarethick] (6.8,0) rectangle ++(\boxwidth,\boxheight);
    \end{tikzpicture}
       \captionsetup{type=figure}   
    
        \captionof{figure}[TPL Alignment: Baseball Pitch]{Examples from  Penn Action dataset (baseball pitch action), showing five equally spaced frames before and after synchronization. We highlight mismatches in the original videos in red, and TPL matching in green. Alignment via TPL successfully synchronizes the key phases of the baseball pitch action.}
    \label{fig:penn_action:baseball_pitch}
\end{figure}

\begin{figure}[H]
    \centering

    % ------------------------------
    % Squat Original (Unsynced)
    % ------------------------------
    \begin{tikzpicture}
    \node[anchor=south west, inner sep=0] (imgA) at (0,0)
    {\includegraphics[trim=0mm 0mm 0mm 0mm,  clip, width=0.98\linewidth]{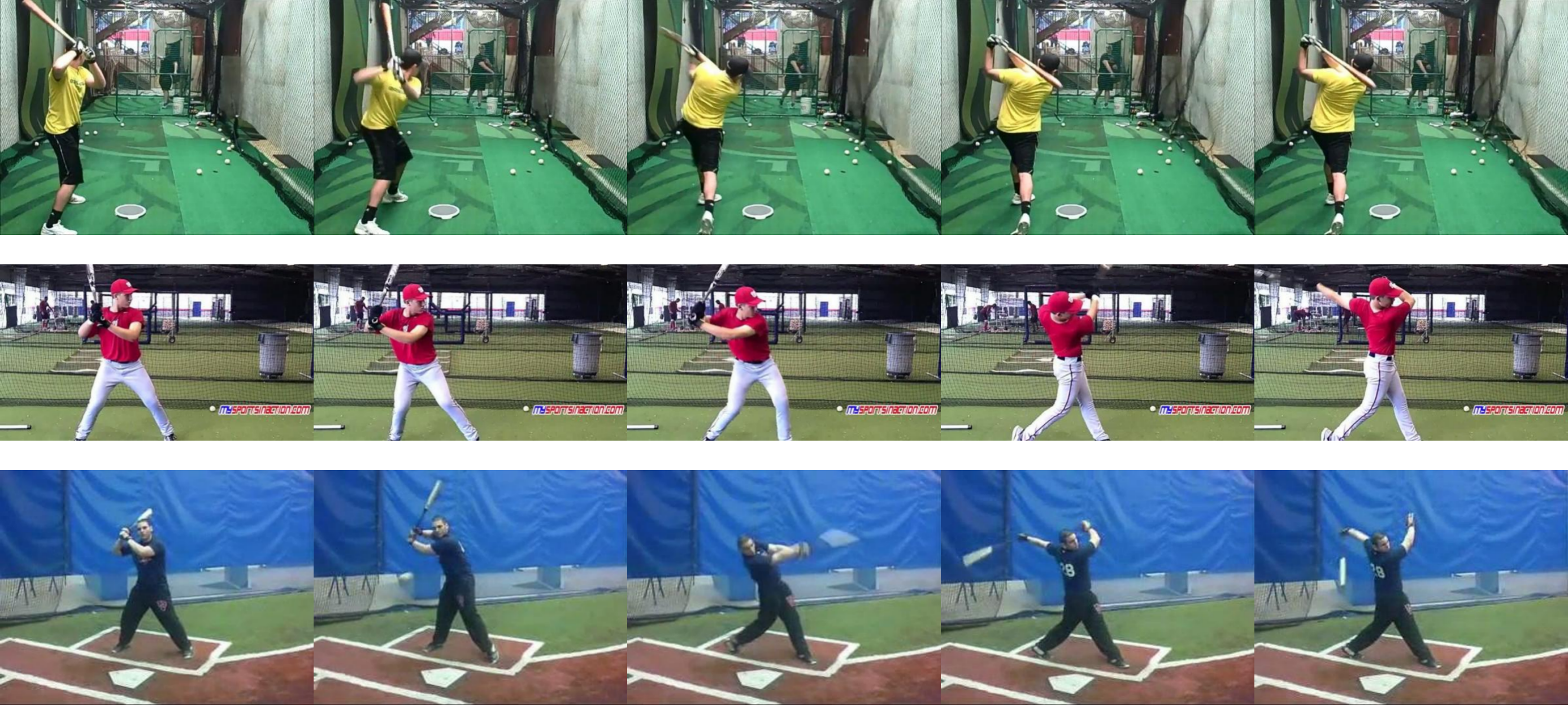}};
    
        \draw[red, line width=\squarethick] (6.8,5.2) rectangle ++(\boxwidth,\boxheight);

        % Overlay squares

        \draw[red, line width=\squarethick] (10.2,5.2-\boxheight) rectangle ++(\boxwidth,\boxheight);
        \draw[red, line width=\squarethick] (6.8,0) rectangle ++(\boxwidth,\boxheight);
    \end{tikzpicture}
    
\hspace{2cm}
        
    % ------------------------------
    % Bear Synced
    % ------------------------------
    \begin{tikzpicture}
    \node[anchor=south west, inner sep=0] (imgA) at (0,0)
    {\includegraphics[trim=0mm 0mm 0mm 0mm, clip, width=0.98\linewidth]{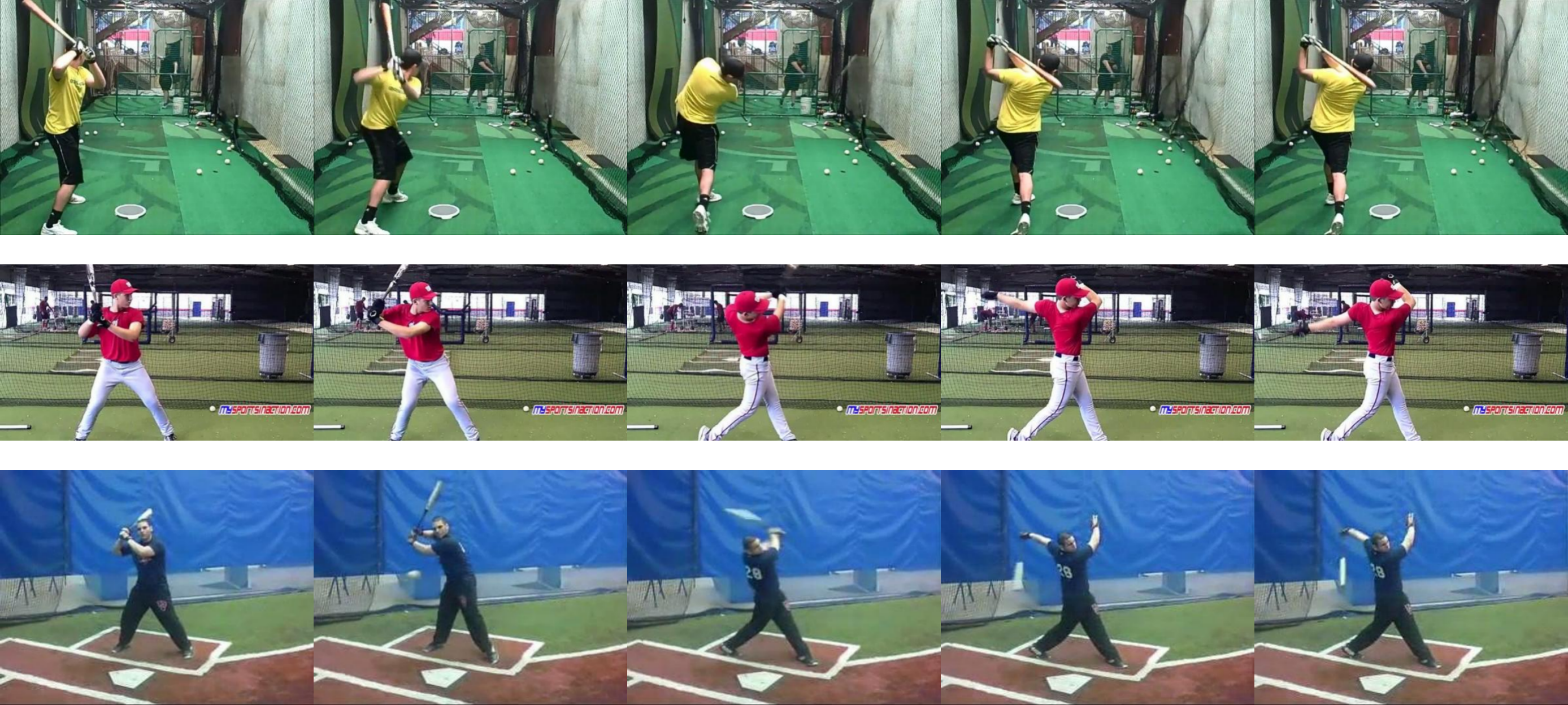}};
        
        % Overlay squares
        \draw[green2, line width=\squarethick] (6.8,5.2) rectangle ++(\boxwidth,\boxheight);
        \draw[green2, line width=\squarethick] (6.8,5.2-\boxheight) rectangle ++(\boxwidth,\boxheight);
        \draw[green2, line width=\squarethick] (6.8,0) rectangle ++(\boxwidth,\boxheight);
    \end{tikzpicture}
       \captionsetup{type=figure}   
    
        \captionof{figure}[TPL Alignment: Baseball Swing]{Examples from  Penn Action dataset (baseball swing action), showing five equally spaced frames before and after synchronization. We highlight mismatches in the original videos in red, and TPL matching in green. Alignment via TPL successfully synchronizes the key phases of the baseball swing action.}
    \label{fig:penn_action:baseball_swing}
\end{figure}

\begin{figure}[H]
    \centering

    % ------------------------------
    % Squat Original (Unsynced)
    % ------------------------------
    \begin{tikzpicture}
    \node[anchor=south west, inner sep=0] (imgA) at (0,0)
    {\includegraphics[trim=0mm 0mm 0mm 0mm,  clip, width=0.98\linewidth]{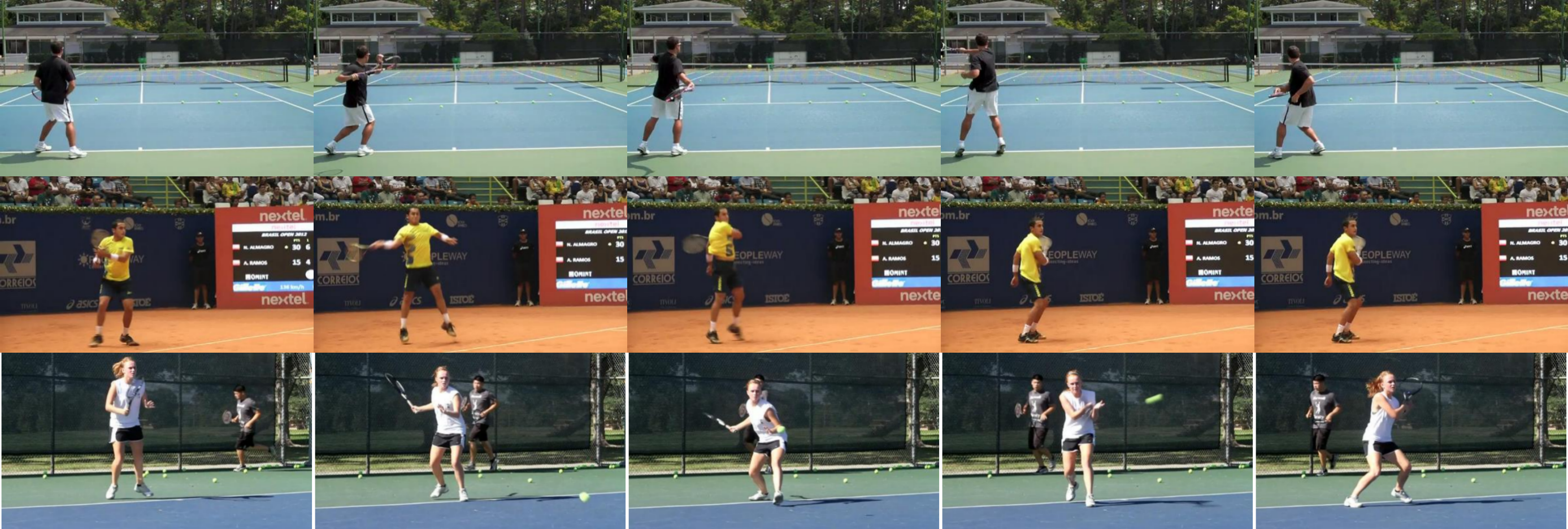}};
    
        \draw[red, line width=\squarethick] (6.8,3.8) rectangle ++(\boxwidth,1.9);

        % Overlay squares

        \draw[red, line width=\squarethick] (3.4,1.9) rectangle ++(\boxwidth,1.9);
        \draw[red, line width=\squarethick] (6.8,0) rectangle ++(\boxwidth,1.9);
    \end{tikzpicture}
    
\hspace{2cm}
        
    % ------------------------------
    % Bear Synced
    % ------------------------------
    \begin{tikzpicture}
    \node[anchor=south west, inner sep=0] (imgA) at (0,0)
    {\includegraphics[trim=0mm 0mm 0mm 0mm, clip, width=0.98\linewidth]{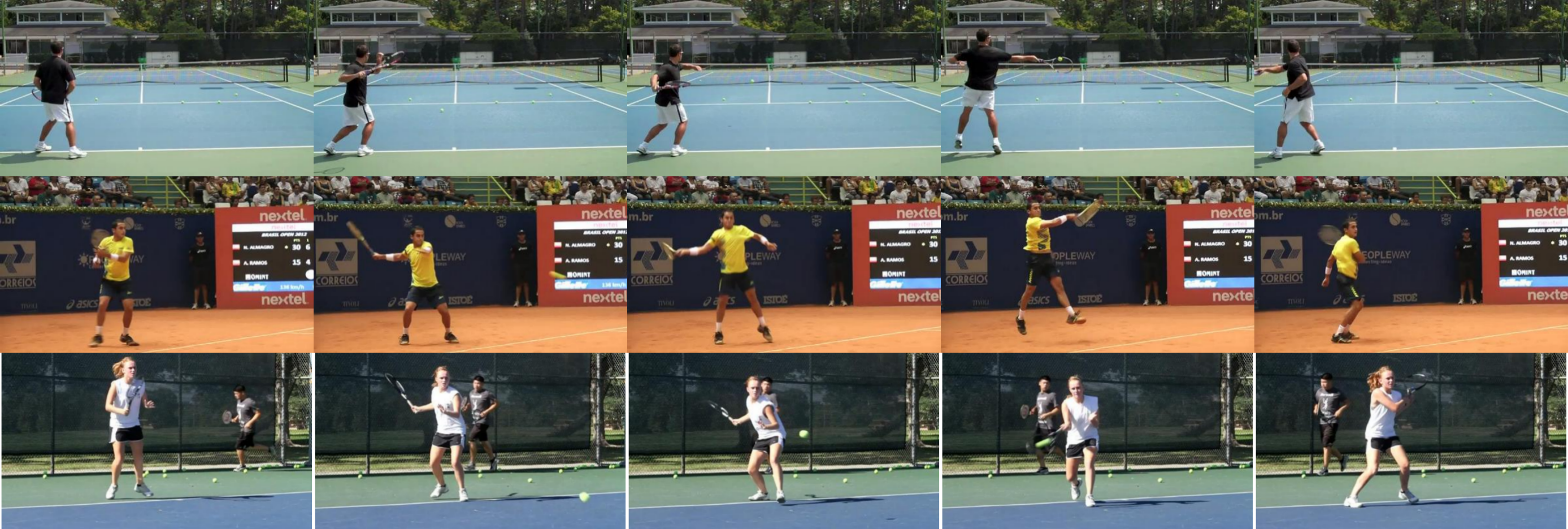}};
        
        % Overlay squares
        \draw[green2, line width=\squarethick] (6.8,3.8) rectangle ++(\boxwidth,1.9);
        \draw[green2, line width=\squarethick] (6.8,1.9) rectangle ++(\boxwidth,1.9);
        \draw[green2, line width=\squarethick] (6.8,0) rectangle ++(\boxwidth,1.9);
    \end{tikzpicture}
       \captionsetup{type=figure}   
    
        \captionof{figure}[TPL Alignment: Tennis Forehand]{Examples from  Penn Action dataset (tennis forehand action), showing five equally spaced frames before and after synchronization. We highlight mismatches in the original videos in red, and TPL matching in green. Alignment via TPL successfully synchronizes the key phases of the tennis forehand action.}
    \label{fig:penn_action:tennis_forehand}
\end{figure}

\clearpage

%% file: supmat/training.tex
\textbf{Diffeomorphic Multitasking Autoencoder (D-MTAE)} modules architecture are presented in~\autoref{tab:encoder_decoder_architecture} and~\autoref{tab:inception_cnn_architecture}.

\subimport{./}{tab_all_archs}

\subsection*{Training details}
For $\psi_{\text{Align}}(\cdot)$, we set the number of cells in the partition of the velocity field to $N_p=16$. We enforce the boundary
condition ($v^{\btheta}[0]=v^{\btheta}[16]=0$) and thus $\text{dim}(\btheta)=15$. We use the InceptionTime backbone for the localization net~\cite{Ismail:2020:inceptiontime} and use the implementation from \texttt{tsai}~\cite{Ignacio:tsai}.
As for the training procedure, we set the batch size to 64 with a learning rate of $10^{-4}$. We jointly train the D-MTAE for all classes for 300 epochs using the AdamW optimizer~\cite{Loshchilo:ICLR:2019:AdamW} with a weight decay of $10^{-4}$. We use a 4090 RTX graphic card for the training of all models.

\subimport{./}{details}

%% file: supmat/tab_all_archs.tex
\begin{table}[ht]
\begin{minipage}[t]{0.5\textwidth}
\centering
 
\subimport{../}{tables/tab_AE}
\end{minipage}
%\hspace{0.005\textwidth}
\begin{minipage}[t]{0.5\textwidth}
\centering
\subimport{../}{tables/tab_dtan_net}
%\scriptsize{GAP -- Global Average Pooling}
\end{minipage}
\end{table}

%% file: tables/tab_AE.tex
%\begin{table}
%\centering
\footnotesize{
\caption{D-MTAE modules architecture.}
\begin{tabular}{c|c|c}
\hline
Operations                  & Output Size & Parameters\\
\hline
\hline
\multicolumn{3}{c}{Encoder network -- $\psi_{\text{encoder}}(\cdot)$} \\
\hline
Conv1d & 128 & [input\_channels, 128, 3, padding=1] \\ 
GELU & 128 & ---\\
Conv1d & 64 & [128, 64, 3, padding=1] \\
GELU & 64 & --- \\
Conv1d & 32 & [64, 32, 3, padding=1] \\
GELU & 32 & --- \\
Conv1d & 16 & [32, 16, 3, padding=1] \\
GELU & 16 & --- \\
Conv1d & 1 & [16, 1, 3, padding=1] \\
\hline
\multicolumn{3}{c}{Decoder network -- $\psi_{\text{decoder}}(\cdot)$} \\
\hline
ConvTranspose1d & 16 & [1, 16, 3, padding=1] \\ 
GELU & 16 & ---\\
ConvTranspose1d & 32 & [16, 32, 3, padding=1] \\
GELU & 32 & --- \\
ConvTranspose1d & 64 & [32, 64, 3, padding=1] \\
GELU & 64 & --- \\
ConvTranspose1d & 128 & [64, 128, 3, padding=1] \\
GELU & 128 & --- \\
ConvTranspose1d & input\_channels & [128, input\_channels, 3, padding=1] \\
\hline
\end{tabular}
\label{tab:encoder_decoder_architecture}
}

%\end{table}

%% file: tables/tab_dtan_net.tex
%\begin{table}
%\centering
\footnotesize{
\caption{Joint alignment network - $\psi_{\text{Align}}(\cdot)$}
\begin{tabular}{c|c|c}
\hline
Operations                  & Output Size & Parameters\\
\hline
\hline
\multicolumn{3}{c}{Inception Block} \\
\hline
Bottleneck Conv      & 32        & [$c$, 1, 32] \\ 
Conv & 32 & [32, 39, 32] \\
Conv & 32 & [32, 19, 32] \\
Conv & 32 & [32, 9, 32] \\
Max Pooling & $c$ & --- \\
Conv      & 32 & [$c$, 1, 32] \\
Concatenation & 128 & --- \\
Batch Norm & 128 & --- \\
ReLU & 128 & --- \\
\hline
\multicolumn{3}{c}{Shortcut} \\
\hline
Conv & 128 & [$c$, 1, 128] \\
Batch Norm & 128 & --- \\
Batch Norm & 128 & --- \\
Addition & 128 & --- \\
ReLU & 128 & --- \\
\hline
\multicolumn{3}{c}{Alignment Head} \\
\hline
GAP & 128 & --- \\
Flatten & 128 & --- \\
Linear Projection & $\text{dim}(\btheta)$ & [128, $\text{dim}(\btheta)$] \\
\hline
\end{tabular}
\label{tab:inception_cnn_architecture}

}

%\end{table}

%% file: supmat/details.tex
\section*{VAE Variant for OpenCLIP and DINO Features}
\label{sec:supmat:vae}
For experiments involving pretrained embeddings from OpenCLIP~\cite{Cherti:CVPR:2023:openclip} and DINO~\cite{Caron:ICCV:2021:dino}, we modify the D-MTAE architecture by replacing the standard autoencoder with a \textbf{Variational Autoencoder (VAE)}.

\paragraph{Latent Sampling.}
The encoder $\Psi_{\text{encoder}}$ now produces a latent distribution per timestep, returning per-frame means $\mu_i \in \RR^{L_i}$ and log-variances $\log \sigma^2_i \in \RR^{L_i}$. The latent trajectory $Z_i \in \RR^{L_i}$ is sampled as:
\begin{align}
    Z_i[t] = \mu_i[t] + \epsilon_i[t] \cdot \sigma_i[t], \quad \epsilon_i[t] \sim \mathcal{N}(0, 1)
\end{align}

\paragraph{Reconstruction Loss.}
After alignment, the decoder reconstructs $\widetilde{U}_i = \Psi_{\text{decoder}}(\widetilde{Z}_i)$ as in the main paper. To handle potential missing or invalid inputs, we use a \textbf{masked reconstruction loss}:
\begin{align}
    \Lcal_{\text{rec}} = \frac{1}{N} \sum_{i=1}^{N} \frac{ \left\| M_i \odot (U_i - \widetilde{U}_i \circ T^{-\btheta_i}) \right\|^2 }{ \sum M_i }
\end{align}
where \( M_i \in \{0,1\}^{C \times L_i} \) is a binary mask and \( \odot \) denotes element-wise multiplication.

\paragraph{KL Divergence.}
To ensure the latent distribution is centered and standardized over time, we apply a \textbf{masked KL divergence loss}:
\begin{align}
    \Lcal_{\text{KL}} = \frac{1}{N} \sum_{i=1}^{N} \frac{ \sum_{t=1}^{L_i} m^{(z)}_i[t] \cdot \left( -\tfrac{1}{2} \big(1 + \log \sigma_i[t]^2 - \mu_i[t]^2 - \sigma_i[t]^2 \big) \right) }{ \sum_t m^{(z)}_i[t] }
\end{align}
where \( m^{(z)}_i[t] \in \{0,1\} \) is a reduced (per-frame) validity mask.

\paragraph{Temporal Smoothness.}
To further regularize the temporal latent trajectories, we penalize sudden changes in \( Z_i \) via a \textbf{smoothness loss}:
\begin{align}
    \Lcal_{\text{smooth}} = \frac{1}{N} \sum_{i=1}^{N} \frac{ \sum_{t=1}^{L_i - 1} m^{(z)}_i[t] \cdot m^{(z)}_i[t+1] \cdot \left( Z_i[t+1] - Z_i[t] \right)^2 }{ \sum_t m^{(z)}_i[t] \cdot m^{(z)}_i[t+1] }
\end{align}
\paragraph{Trajectory Variance Loss.}
To encourage coherent temporal dynamics across sequences in a batch, we introduce a trajectory variance loss. For each sequence, we extract \( Z_i^{\text{sub}} \in \RR^K \) by uniformly sampling \( K \) valid timesteps from the latent trajectory \( Z_i \). The loss penalizes deviation from the batch-wise mean trajectory:
\begin{align}
    \Lcal_{\text{traj-var}} = \frac{1}{N} \sum_{i=1}^{N} \left\| Z_i^{\text{sub}} - \overline{Z}^{\text{sub}} \right\|^2,
\end{align}
where \( \overline{Z}^{\text{sub}} = \frac{1}{N} \sum_{i=1}^{N} Z_i^{\text{sub}} \). This regularization encourages latent trajectories to evolve with similar temporal structure across samples, without enforcing identity or similarity in content.

\paragraph{Final Objective.}
The total training loss used for OpenCLIP/DINO features becomes:
\begin{align}
    \Lcal_{\text{TPL-VAE}} = \lambda_t \Lcal_{\mathrm{ICAE}} + \Lcal_{\text{rec}} + \beta \cdot \Lcal_{\text{KL}} + \gamma \cdot \Lcal_{\text{smooth}} + \alpha \cdot \Lcal_{\text{traj-var}}
\end{align}
where \( \lambda_t \) is the annealed ICAE weight, and \( \beta, \gamma, \alpha \) are fixed hyperparameters.

\paragraph{Notes.} 
- When using this variant, only the encoder and decoder are changed; the alignment module $\Psi_{\text{Align}}$ and CPAB transformations remain as described in the main paper.
- This change is only applied to experiments where input features are obtained from pretrained OpenCLIP or DINO models.

%% file: paper.bbl
\begin{thebibliography}{35}
\providecommand{\natexlab}[1]{#1}
\providecommand{\url}[1]{\texttt{#1}}
\expandafter\ifx\csname urlstyle\endcsname\relax
  \providecommand{\doi}[1]{doi: #1}\else
  \providecommand{\doi}{doi: \begingroup \urlstyle{rm}\Url}\fi

\bibitem[Caron et~al.(2021)Caron, Touvron, Misra, J{\'e}gou, Mairal, Bojanowski, and Joulin]{Caron:ICCV:2021:dino}
Mathilde Caron, Hugo Touvron, Ishan Misra, Herv{\'e} J{\'e}gou, Julien Mairal, Piotr Bojanowski, and Armand Joulin.
\newblock Emerging properties in self-supervised vision transformers.
\newblock In \emph{Proceedings of the IEEE/CVF international conference on computer vision}, pages 9650--9660, 2021.

\bibitem[Chen et~al.(2022)Chen, Wei, Li, and Cai]{Chen:CVPR:2022:CARL}
Minghao Chen, Fangyun Wei, Chong Li, and Deng Cai.
\newblock Frame-wise action representations for long videos via sequence contrastive learning.
\newblock In \emph{CVPR}, 2022.

\bibitem[Cherti et~al.(2023)Cherti, Beaumont, Wightman, Wortsman, Ilharco, Gordon, Schuhmann, Schmidt, and Jitsev]{Cherti:CVPR:2023:openclip}
Mehdi Cherti, Romain Beaumont, Ross Wightman, Mitchell Wortsman, Gabriel Ilharco, Cade Gordon, Christoph Schuhmann, Ludwig Schmidt, and Jenia Jitsev.
\newblock Reproducible scaling laws for contrastive language-image learning.
\newblock In \emph{Proceedings of the IEEE/CVF conference on computer vision and pattern recognition}, pages 2818--2829, 2023.

\bibitem[Cuturi and Blondel(2017)]{Cuturi:2017:soft}
Marco Cuturi and Mathieu Blondel.
\newblock Soft-dtw: a differentiable loss function for time-series.
\newblock \emph{arXiv preprint arXiv:1703.01541}, 2017.

\bibitem[Dau et~al.(2019)Dau, Bagnall, Kamgar, Yeh, Zhu, Gharghabi, Ratanamahatana, and Keogh]{Dau:2019:ucr}
Hoang~Anh Dau, Anthony Bagnall, Kaveh Kamgar, Chin-Chia~Michael Yeh, Yan Zhu, Shaghayegh Gharghabi, Chotirat~Ann Ratanamahatana, and Eamonn Keogh.
\newblock The ucr time series archive.
\newblock \emph{IEEE/CAA Journal of Automatica Sinica}, 2019.

\bibitem[Dave et~al.(2024)Dave, Heilbron, Shah, and Jenni]{Dave:ECCV:2024:sync}
Ishan~Rajendrakumar Dave, Fabian~Caba Heilbron, Mubarak Shah, and Simon Jenni.
\newblock Sync from the sea: retrieving alignable videos from large-scale datasets.
\newblock In \emph{European Conference on Computer Vision}, pages 371--388. Springer, 2024.

\bibitem[Dong et~al.(2020)Dong, Shuai, Zhang, Liu, Zhou, and Bao]{Dong:ECCV:2020:NetVideos}
Junting Dong, Qing Shuai, Yuanqing Zhang, Xian Liu, Xiaowei Zhou, and Hujun Bao.
\newblock Motion capture from internet videos.
\newblock In \emph{Computer Vision--ECCV 2020: 16th European Conference, Glasgow, UK, August 23--28, 2020, Proceedings, Part II 16}, pages 210--227. Springer, 2020.

\bibitem[Dwibedi et~al.(2019)Dwibedi, Aytar, Tompson, Sermanet, and Zisserman]{Debidatta:CVPR:2019:TCCL}
Debidatta Dwibedi, Yusuf Aytar, Jonathan Tompson, Pierre Sermanet, and Andrew Zisserman.
\newblock Temporal cycle-consistency learning.
\newblock In \emph{CVPR}, 2019.

\bibitem[Freifeld et~al.(2015)Freifeld, Hauberg, Batmanghelich, and Fisher~III]{Freifeld:ICCV:2015:CPAB}
Oren Freifeld, S{\o}ren Hauberg, Kayhan Batmanghelich, and John~W. Fisher~III.
\newblock Highly-expressive spaces of well-behaved transformations: Keeping it simple.
\newblock In \emph{ICCV}, 2015.

\bibitem[Freifeld et~al.(2017)Freifeld, Hauberg, Batmanghelich, and Fisher~III]{Freifeld:PAMI:2017:CPAB}
Oren Freifeld, S{\o}ren Hauberg, Kayhan Batmanghelich, and John~W. Fisher~III.
\newblock Transformations based on continuous piecewise-affine velocity fields.
\newblock \emph{IEEE TPAMI}, 2017.

\bibitem[Hadji et~al.(2021)Hadji, Derpanis, and Jepson]{Hadji:CVPR:2021:gta}
Isma Hadji, Konstantinos~G Derpanis, and Allan~D Jepson.
\newblock Representation learning via global temporal alignment and cycle-consistency.
\newblock In \emph{CVPR}, 2021.

\bibitem[Haresh et~al.(2021)Haresh, Kumar, Coskun, Syed, Konin, Zia, and Tran]{Haresh:CVPR:2021:LAV}
Sanjay Haresh, Sateesh Kumar, Huseyin Coskun, Shahram~N Syed, Andrey Konin, Zeeshan Zia, and Quoc-Huy Tran.
\newblock Learning by aligning videos in time.
\newblock In \emph{CVPR}, 2021.

\bibitem[Ismail~Fawaz et~al.(2020)Ismail~Fawaz, Lucas, Forestier, Pelletier, Schmidt, Weber, Webb, Idoumghar, Muller, and Petitjean]{Ismail:2020:inceptiontime}
Hassan Ismail~Fawaz, Benjamin Lucas, Germain Forestier, Charlotte Pelletier, Daniel~F Schmidt, Jonathan Weber, Geoffrey~I Webb, Lhassane Idoumghar, Pierre-Alain Muller, and Fran{\c{c}}ois Petitjean.
\newblock Inceptiontime: Finding alexnet for time series classification.
\newblock \emph{Data Mining and Knowledge Discovery}, 2020.

\bibitem[Liang et~al.(2017)Liang, Huang, Chen, and Hauptmann]{Liang:ICASSP:2017:synchronization}
Junwei Liang, Poyao Huang, Jia Chen, and Alexander Hauptmann.
\newblock Synchronization for multi-perspective videos in the wild.
\newblock In \emph{2017 IEEE International Conference on Acoustics, Speech and Signal Processing (ICASSP)}, pages 1592--1596. IEEE, 2017.

\bibitem[Liu et~al.(2022)Liu, Tekin, Coskun, Vineet, Fua, and Pollefeys]{Liu:CVPR:2022:VAVA}
Weizhe Liu, Bugra Tekin, Huseyin Coskun, Vibhav Vineet, Pascal Fua, and Marc Pollefeys.
\newblock Learning to align sequential actions in the wild.
\newblock In \emph{CVPR}, 2022.

\bibitem[Loshchilov and Hutter(2019)]{Loshchilo:ICLR:2019:AdamW}
Ilya Loshchilov and Frank Hutter.
\newblock Decoupled weight decay regularization.
\newblock In \emph{International Conference on Learning Representations}, 2019.

\bibitem[Martinez et~al.(2022)Martinez, Viles, and Olaizola]{Martinez:ICML:2022:closed}
I{\~n}igo Martinez, Elisabeth Viles, and Igor~G Olaizola.
\newblock Closed-form diffeomorphic transformations for time series alignment.
\newblock In \emph{ICML}. PMLR, 2022.

\bibitem[Oguiza(2022)]{Ignacio:tsai}
Ignacio Oguiza.
\newblock tsai - a state-of-the-art deep learning library for time series and sequential data.
\newblock Github, 2022.

\bibitem[Oquab et~al.(2023)Oquab, Darcet, Moutakanni, Vo, Szafraniec, Khalidov, Fernandez, Haziza, Massa, El-Nouby, et~al.]{Oquab:2023:dinov2}
Maxime Oquab, Timoth{\'e}e Darcet, Th{\'e}o Moutakanni, Huy Vo, Marc Szafraniec, Vasil Khalidov, Pierre Fernandez, Daniel Haziza, Francisco Massa, Alaaeldin El-Nouby, et~al.
\newblock Dinov2: Learning robust visual features without supervision.
\newblock \emph{arXiv preprint arXiv:2304.07193}, 2023.

\bibitem[Paszke et~al.(2019)Paszke, Gross, Massa, Lerer, Bradbury, Chanan, Killeen, Lin, Gimelshein, Antiga, et~al.]{Paszke:NIPS:2019:pytorch}
Adam Paszke, Sam Gross, Francisco Massa, Adam Lerer, James Bradbury, Gregory Chanan, Trevor Killeen, Zeming Lin, Natalia Gimelshein, Luca Antiga, et~al.
\newblock Pytorch: An imperative style, high-performance deep learning library.
\newblock \emph{NeurIPS}, 2019.

\bibitem[Petitjean et~al.(2011)Petitjean, Ketterlin, and Gan{\c{c}}arski]{Petitjean:2011:global}
Fran{\c{c}}ois Petitjean, Alain Ketterlin, and Pierre Gan{\c{c}}arski.
\newblock A global averaging method for dynamic time warping, with applications to clustering.
\newblock \emph{Pattern Recognition}, 2011.

\bibitem[Petitjean et~al.(2014)Petitjean, Forestier, Webb, Nicholson, Chen, and Keogh]{Petitjean:2014:dynamic}
Fran{\c{c}}ois Petitjean, Germain Forestier, Geoffrey~I Webb, Ann~E Nicholson, Yanping Chen, and Eamonn Keogh.
\newblock Dynamic time warping averaging of time series allows faster and more accurate classification.
\newblock In \emph{IEEE ICDM}, 2014.

\bibitem[Radford et~al.(2021)Radford, Kim, Hallacy, Ramesh, Goh, Agarwal, Sastry, Askell, Mishkin, Clark, et~al.]{Radford:ICML:2021:CLIP}
Alec Radford, Jong~Wook Kim, Chris Hallacy, Aditya Ramesh, Gabriel Goh, Sandhini Agarwal, Girish Sastry, Amanda Askell, Pamela Mishkin, Jack Clark, et~al.
\newblock Learning transferable visual models from natural language supervision.
\newblock In \emph{International conference on machine learning}, pages 8748--8763. PmLR, 2021.

\bibitem[Sakoe(1971)]{Sakoe:ICA:1971:DTW1}
H. Sakoe.
\newblock Dynamic-programming approach to continuous speech recognition.
\newblock \emph{The International Congress of Acoustics}, 1971.

\bibitem[Sakoe and Chiba(1978)]{Sakoe:ASSP:1971:DTW2}
H. Sakoe and S. Chiba.
\newblock Dynamic programming algorithm optimization for spoken word recognition.
\newblock \emph{IEEE TASSP}, 1978.

\bibitem[Sermanet et~al.(2017)Sermanet, Lynch, Hsu, and Levine]{DBLP:journals/corr/SermanetLHL17}
Pierre Sermanet, Corey Lynch, Jasmine Hsu, and Sergey Levine.
\newblock Time-contrastive networks: Self-supervised learning from multi-view observation.
\newblock \emph{CoRR}, abs/1704.06888, 2017.

\bibitem[Sermanet et~al.(2018)Sermanet, Lynch, Chebotar, Hsu, Jang, Schaal, Levine, and Brain]{Sermanet:ICRA:2018:pouring}
Pierre Sermanet, Corey Lynch, Yevgen Chebotar, Jasmine Hsu, Eric Jang, Stefan Schaal, Sergey Levine, and Google Brain.
\newblock Time-contrastive networks: Self-supervised learning from video.
\newblock In \emph{ICRA}. IEEE, 2018.

\bibitem[Shapira~Weber and Freifeld(2023)]{Shapira:ICML:2023:rfdtan}
Ron Shapira~Weber and Oren Freifeld.
\newblock Regularization-free diffeomorphic temporal alignment nets.
\newblock In \emph{ICML}. PMLR, 2023.

\bibitem[Shapira~Weber et~al.(2019)Shapira~Weber, Eyal, Skafte~Detlefsen, Shriki, and Freifeld]{Shapira:NIPS:2019:DTAN}
Ron Shapira~Weber, Matan Eyal, Nicki Skafte~Detlefsen, Oren Shriki, and Oren Freifeld.
\newblock Diffeomorphic temporal alignment nets.
\newblock In \emph{NeurIPS}, 2019.

\bibitem[Shrstha et~al.(2007)Shrstha, Barbieri, and Weda]{Shrstha:ACM:2007:synchronization}
Prarthana Shrstha, Mauro Barbieri, and Hans Weda.
\newblock Synchronization of multi-camera video recordings based on audio.
\newblock In \emph{Proceedings of the 15th ACM international conference on Multimedia}, pages 545--548, 2007.

\bibitem[Snell et~al.(2017)Snell, Swersky, and Zemel]{Snell:NIPS:2017:prototypical}
Jake Snell, Kevin Swersky, and Richard Zemel.
\newblock Prototypical networks for few-shot learning.
\newblock \emph{NeurIPS}, 2017.

\bibitem[Tong et~al.(2022)Tong, Song, Wang, and Wang]{Tong:NeurIPS:2022:videomae}
Zhan Tong, Yibing Song, Jue Wang, and Limin Wang.
\newblock Videomae: Masked autoencoders are data-efficient learners for self-supervised video pre-training.
\newblock \emph{Advances in neural information processing systems}, 35:\penalty0 10078--10093, 2022.

\bibitem[Wang et~al.(2014)Wang, Schroers, Zimmer, Gross, and Sorkine-Hornung]{Wang:ACM:2014:videosnapping}
Oliver Wang, Christopher Schroers, Henning Zimmer, Markus Gross, and Alexander Sorkine-Hornung.
\newblock Videosnapping: Interactive synchronization of multiple videos.
\newblock \emph{ACM Transactions on Graphics (TOG)}, 33\penalty0 (4):\penalty0 1--10, 2014.

\bibitem[Zhang et~al.(2023)Zhang, Liu, Zheng, and Su]{zhang2023modeling}
Heng Zhang, Daqing Liu, Qi Zheng, and Bing Su.
\newblock Modeling video as stochastic processes for fine-grained video representation learning.
\newblock In \emph{CVPR}, 2023.

\bibitem[Zhang et~al.(2013)Zhang, Zhu, and Derpanis]{zhang:ICCV:2013:penn}
Weiyu Zhang, Menglong Zhu, and Konstantinos~G Derpanis.
\newblock From actemes to action: A strongly-supervised representation for detailed action understanding.
\newblock In \emph{ICCV}, 2013.

\end{thebibliography}
